\documentclass[journal]{IEEEtran}

\usepackage{graphicx}%
\usepackage{multirow}%
\usepackage{amsmath,amssymb,amsfonts}%
\usepackage{amsthm}%
\usepackage{mathrsfs}%
\usepackage{xcolor}%
\usepackage{textcomp}%
\usepackage{manyfoot}%
\usepackage{booktabs}%
\usepackage{algorithm}%
\usepackage{algorithmicx}%
\usepackage{algpseudocode}%
\usepackage{listings}%
\usepackage{soul}

\usepackage{booktabs,makecell}
\usepackage{supertabular} 
\usepackage{lipsum}       
\usepackage{longtable}
\usepackage{tabularx}
\usepackage{caption}
\usepackage{siunitx}

\begin{document}
\bstctlcite{IEEEexample:BSTcontrol}

\title{Deep Reinforcement Learning for Robotic Bipedal Locomotion: A Brief Survey}
%





%

\author{Lingfan~Bao$^{1}$, Joseph~Humphreys$^{12}$, Tianhu~Peng$^{1}$ and Chengxu~Zhou$^{1}$
\thanks{This work was partially supported by the Royal Society [grant number RG\textbackslash R2\textbackslash232409] and the Advanced Research and Invention Agency [grant number SMRB-SE01-P06].}
\thanks{$^{1}$Department of Computer Science, University College London, UK. {\tt\footnotesize chengxu.zhou@ucl.ac.uk}}
\thanks{$^{2}$ School of Mechanical Engineering, University of Leeds, UK.}
}
\maketitle

\begin{abstract}
Bipedal robots are gaining global recognition due to their potential applications and the rapid advancements in artificial intelligence, particularly through Deep Reinforcement Learning (DRL). While DRL has significantly advanced bipedal locomotion, the development of a unified framework capable of handling a wide range of tasks remains an ongoing challenge. This survey systematically categorises, compares, and analyses existing DRL frameworks for bipedal locomotion, organising them into end-to-end and hierarchical control schemes. End-to-end frameworks are evaluated based on their learning approaches, whereas hierarchical frameworks are examined in terms of their layered structures that integrate learning-based and traditional model-based methods. We provide a detailed evaluation of the composition, strengths, limitations, and capabilities of each framework. Furthermore, this survey identifies key research gaps and proposes future directions aimed at creating a more integrated and efficient unified framework for bipedal locomotion, with broad applicability in real-world environments.
\end{abstract}


%
\IEEEpeerreviewmaketitle

\section{Introduction}\label{Sec:Introduction}
Humans navigate complex environments and perform diverse locomotion tasks with remarkable efficiency using only two legs. Bipedal robots, which closely mimic the human form, possess distinct advantages over wheeled or tracked alternatives, particularly when traversing uneven and challenging terrains. Furthermore, bipedal humanoid robots are specifically designed to operate in human-centric environments, enabling seamless interaction with tools and infrastructure intended for human use. This makes them highly adaptable to a wide range of tasks in such settings.

\begin{figure}
    \centering
    \includegraphics[trim={3.5cm 7.3cm 3.6cm 7.2cm},clip,width=\linewidth]{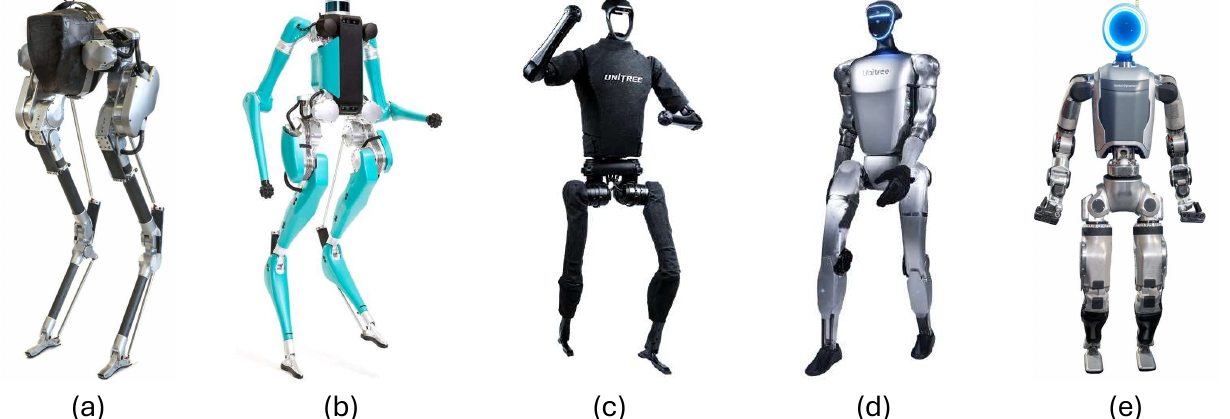}
    \vspace{-3mm}
    \caption{Representative bipedal and humanoid robots illustrating the diversity of platforms for locomotion research and development. (a) Cassie: a torque-controlled bipedal robot designed for agile locomotion. (b) Digit: a full-sized humanoid robot evolved from Cassie and actuated by torque control. (c) H1: a full-size, electric, torque-controlled humanoid robot developed by Unitree Robotics. (d) G1: a compact humanoid robot from Unitree featuring lightweight design and high joint backdrivability. (e) Atlas: a fully electric humanoid robot developed by Boston Dynamics.}
    \label{fig:bipeds}
\end{figure} 

As a result, bipedal robots hold significant potential for real-world applications \cite{2024_tong_advancements_review}. In manufacturing, they can perform tasks efficiently without requiring additional tools, thereby enhancing productivity and reducing labour demands \cite{industrial_robotics_applications, collaborative_industrial, 2023_vidoeo_demo_digit_locomaniputlation}. Their agility is particularly advantageous in complex environments such as multi-level workplaces. Bipedal robots are also well suited to tasks that involve the use of human-designed tools, making them valuable for assisting in daily activities, healthcare, and rehabilitation \cite{2019_bingjing_human_exoskeleton}. Moreover, they show considerable promise in search-and-rescue operations, where they can navigate hazardous and unpredictable terrains \cite{2015_underwaterrobot_review,quadSpace1,quadSpace2}.

Traditional approaches to bipedal locomotion control, such as model-based methods, have been prevalent since the 1980s \cite{2017_biped_model-based_method_review,2021_Jenna_bipedrobot_review,2021_Justin_review_recentporgressinleggedrobot}. Early methods, such as the Linear Inverted Pendulum Model (LIPM) \cite{2013_3DSLIPmodel_running}, provided simplified representations of the dynamics involved in bipedal motion, enabling easier analysis and control. As research progressed, full dynamic models were introduced to better capture the complexities of real-world locomotion. Advanced methods such as Model Predictive Control (MPC) \cite{2022_Hou_drlasparametertoMPC,2023_junheng_steppingstones_MPCWBC} and Trajectory Optimisation (TO) \cite{2016_LIPM_trajectoryoptimization,2019_tianyu_ATRIAS_PPO_joint0level_trajectory_learnedhighlevelpolicy,2017_biped_model-based_method_review} exploit predefined dynamic models to solve constrained optimal-control problems that plan footsteps, centre-of-mass (CoM) motion, and contact forces. While model-based approaches offer rapid convergence and predictive capabilities, they often struggle in dynamically complex and uncertain environments where adaptability is essential.

Reinforcement learning (RL)-based methods, particularly deep reinforcement learning (DRL), are effective in optimising robot control policies through direct interaction with the environment \cite{2020_systematic_review}, which provides a distinct advantage. Unlike model-based approaches, which rely on predefined dynamics and may fail under unforeseen conditions, DRL enables robots to autonomously discover control strategies through trial and error, achieving greater adaptability and robustness in diverse environments. In addition, hybrid methods that combine model-based and learning-based techniques further enhance planning and control by leveraging the strengths of both paradigms.

Despite these advancements, research in DRL-based locomotion remains highly fragmented, with inconsistencies in training pipelines, reward formulations, observation spaces, and evaluation setups that hinder systematic benchmarking and slow progress towards generalisable locomotion capabilities. Moreover, many methods are tailored to specific morphologies or tasks, offering limited transferability across embodiments and environments.

This fragmentation motivates the following central research questions: \textit{To what extent has current research achieved generalisation and robustness across diverse morphologies, terrains, and locomotion tasks? If full generalisation has not yet been realised, how can existing DRL approaches be organised and extended towards a unified framework that enables such capability in bipedal robots?} In this context, the present survey seeks to categorise emerging DRL frameworks for bipedal locomotion, identify their key limitations, and outline opportunities for integration and convergence towards unification.

To address these aspects, we first clarify the ultimate goal of robot learning: to develop systems that exhibit generalisation, adaptability, and robustness across diverse morphologies, tasks, and environments. The unified framework is therefore not the final destination but a conceptual scaffold emerging from the consolidation of current DRL research efforts. Its role is to organise fragmented methodologies through shared interfaces, training conventions, and evaluation protocols, thereby promoting steady progress towards the broader goal of generalisable and adaptive robot-learning systems.

Guided by these definitions, this survey examines recent advancements in DRL-based frameworks, categorising control schemes into two primary types: (i) end-to-end and (ii) hierarchical. End-to-end frameworks directly map robot states to joint-level control outputs, while hierarchical frameworks decompose decision-making into multiple layers. In hierarchical systems, a High-Level (HL) planner governs navigation and path planning, while a Low-Level (LL) controller handles fundamental locomotion tasks. The task-level decision-making tier interfaces directly with user commands or predefined tasks, forming a structured approach to robotic control.

The evolution of RL in bipedal robotics has largely advanced through the end-to-end learning paradigm. Early studies in 2004 applied simple policy-gradient methods to 2D bipeds \cite{2004_Tedrake_RL_bipedal_policygradient,2004_model-based_RL_learnpoincaremap_simple_bipedrobot}, while later breakthroughs in DRL enabled policy training in high-fidelity physics simulators \cite{2017_Xuebinpeng_animation_deeploco_DRL_hiarachysystem_referencemotion_deepconvolutionalneuralnewtwork_jointangle_soccer.,2018_deepmimic_animation_PPO_deepmimic_desrieddirection_physics-based_animation_ATLAS_task,2018_wenhao_yu_DRL_withoutpredefine_symmetrygait_PPO_jointanglePD_}. As robotic hardware matured, an increasing variety of bipedal and humanoid platforms emerged, supporting extensive evaluation of DRL-based locomotion across diverse morphologies, as illustrated in Fig.~\ref{fig:bipeds}. This evolution marked the transition from purely simulation-based training to simulation-to-real (sim-to-real) transfer, where policies trained in simulators are deployed on physical robots. In 2020, the first successful sim-to-real transfer of an end-to-end DRL locomotion policy was achieved on the 3D torque-controlled bipedal robot Cassie \cite{2020_Xie_firstsim2real_}. Subsequent work explored two principal learning paradigms: reference-based learning, which leverages TO-generated data or motion-capture data to guide policy training \cite{2021_michael_velocity-basedcontrol_motioncapturedata,2024_speratebody_expressive_allmotion,2024_tang_humanmimic_adversarial,2024_Zhang_wholebody_adversarial_motion_priors}; and reference-free learning, where policies are trained entirely from scratch to autonomously discover control strategies \cite{2021_siekmann_sim2real_nonreference_perodicreward_DRL_e2e_LSTM_PPO_cassie}. These developments demonstrate that end-to-end frameworks can achieve robust and versatile locomotion skills across complex terrains and dynamic environments \cite{2022_OSUDRL_steppingstone_referencefree_predictionfeasiblefootsteps_camera_benchmark,2023_Chongben_jumping_DDPG_multi-task_network_simulationonly,2024_li_ucb_unifiedframework_PPO_IOput_teacherstudentcompare_IOhistory_e2etraining}.

\begin{figure*}
    \centering
    \includegraphics[trim={1cm 0.4cm 0cm 0.7cm},clip,width=\textwidth]{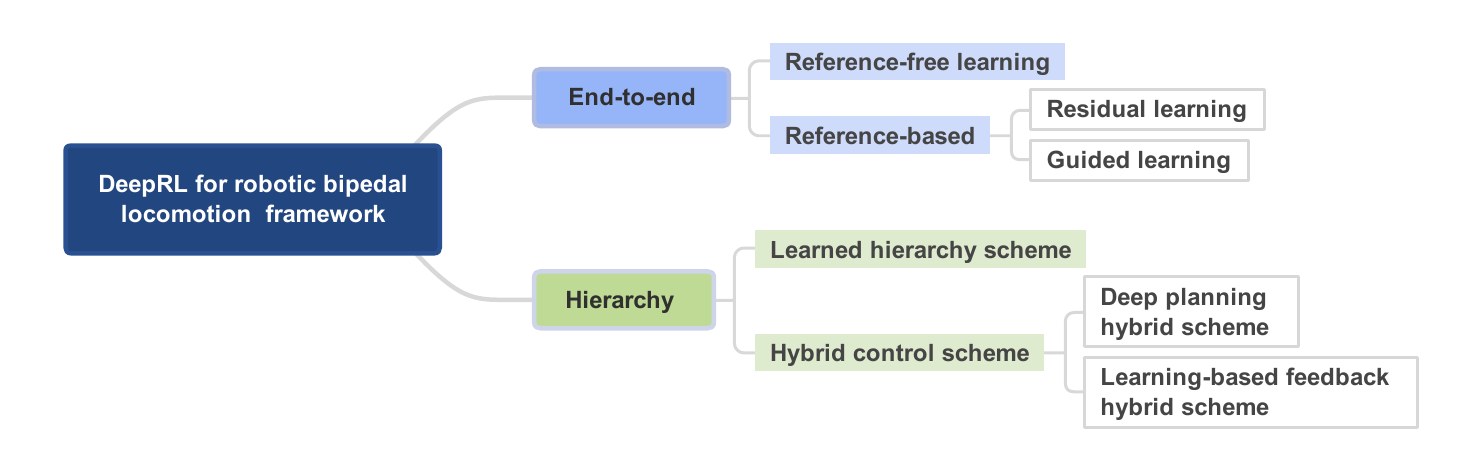}
    \caption{Classification of DRL-based control schemes. The approaches are broadly categorised into two main paradigms: end-to-end frameworks, which learn a single policy from sensory inputs to motor commands; and hierarchical frameworks, which decompose the control problem into multiple levels. Within the end-to-end paradigm, a key distinction is drawn between reference-free learning (learning from scratch) and reference-based learning (tracking a predefined motion). Hierarchical structures include hybrid control schemes, which synergistically combine learned components with traditional model-based controllers.}
    \label{fig:framework_catalog}
\end{figure*}

Similarly, hierarchical structures have garnered significant interest. Within this subset, the hybrid approach combines RL-based and model-based methods to enhance both planning and control strategies. Hybrid architectures often integrate learning-based and model-based modules to combine adaptability with physical consistency. One representative design couples a learned HL planner with an LL model-based controller, forming a cascade-structure or deep-planning hybrid scheme \cite{2019_tianyu_ATRIAS_PPO_joint0level_trajectory_learnedhighlevelpolicy,2021_duan_DRL_task-spaceaction_hiarachycotnrolscheme_inversedynamiccontroller,2022_DRL_cascade_motionplanningpolicy_HZD_torso_PPOwithRNN}. Alternatively, DRL feedback-control hybrids embed learned control policies within model-based feedback loops to enhance tracking precision and disturbance rejection \cite{2022_Singh_humanoid_e2erlwithfootplannar_hierarchyscheme_PPO_simulation_HRP-5P,2023_wangsong_DRL_footsteptrakcing_hybridscheme_experiment_PPO_stair_task_LIPM_curriculumlearning}. Learned hierarchical control schemes \cite{2017_DRL_survey} decompose locomotion into multiple layers, each focusing on specific functions such as navigation and fundamental locomotion skills \cite{2017_Xuebinpeng_animation_deeploco_DRL_hiarachysystem_referencemotion_deepconvolutionalneuralnewtwork_jointangle_soccer.,2018_deepmimic_animation_PPO_deepmimic_desrieddirection_physics-based_animation_ATLAS_task,2023_Wei_learned_hierarchy_framework_wheeled_bipedalrobot}. To provide a clearer overview of the current landscape, we categorise existing DRL frameworks as shown in Fig.~\ref{fig:framework_catalog}.


Current progress across both end-to-end and hierarchical paradigms indicates that a unified framework for DRL-based bipedal locomotion is still far from being realised. Establishing such a framework is essential for consolidating diverse learning pipelines, standardising evaluation metrics, and enabling transferable locomotion capabilities across different robot morphologies. As locomotion tasks become increasingly complex, ranging from basic stabilisation to dynamic parkour and loco-manipulation, the need for consistent benchmarking has intensified.The DARPA Robotics Challenge exemplified this trend by introducing one of the first large-scale evaluation platforms for bipedal humanoids performing real-world locomotion and manipulation tasks \cite{2015_DAPRA_Challenge_1}, highlighting the importance of robustness and practical deployment.

Although several reviews discuss RL for general robotics \cite{2020_systematic_review} and model-based methods for bipedal robots \cite{2017_biped_model-based_method_review,2021_Jenna_bipedrobot_review,2021_Justin_review_recentporgressinleggedrobot}, none specifically focus on DRL-based frameworks for bipeds. To address this gap, this survey reviews relevant literature according to the following selection criteria: (1) studies that investigate DRL frameworks specifically for bipedal robots; (2) research involving both simulated and physical bipedal robots; (3) approaches that improve policy transfer from simulation to real-world environments; and (4) publications from the last five years (2018–April 2024) sourced from reputable databases and conferences, including Google Scholar, IEEE Xplore, Web of Science, arXiv, and major robotics venues such as CoRL, RSS, ICRA, IROS, and Humanoids.

The search was conducted using the key terms ``deep reinforcement learning'' or ``reinforcement learning'' in combination with ``bipedal locomotion'', ``bipedal walking'', ``biped robot'', ``humanoid robot'', or ``legged robot''. The most relevant and impactful works were manually selected for further review. This survey is intended for readers with a foundational background in robotics who are transitioning to DRL methods, providing an overview of a wide range of approaches with simplified explanations where appropriate. For clarity, throughout this paper the term ``humanoid robot'' refers specifically to anthropomorphic bipedal robots.

The primary contributions of this survey are:
\begin{itemize}
    \item A comprehensive summary and cataloguing of DRL-based frameworks for bipedal locomotion.
    \item A detailed comparison of each control scheme, highlighting their strengths, limitations, and distinctive characteristics.
    \item The identification of current challenges and the provision of insightful future research directions.
\end{itemize}

The survey is organised as follows: Section~\ref{Sec:e2e_framework} discusses end-to-end frameworks, categorised by learning approaches; Section~\ref{Sec:Hierarchy_framework} presents hierarchical frameworks, classified into three main types; Section~\ref{Sec:limitations_challenges} outlines key limitations and challenges, linking them to the preceding discussions; Section~\ref{Sec:future_directions_opportunities} explores potential pathways, opportunities, and two proposed conceptual models that extend the end-to-end and hierarchical paradigms; finally, Section~\ref{sec:conclusion} concludes the survey.

\section{End-to-end framework}\label{Sec:e2e_framework}
The end-to-end DRL framework represents a holistic approach in which a single neural network (NN) policy, denoted \(\pi(\cdot) : \mathcal{X} \rightarrow \mathcal{U}\), directly maps sensory inputs \(\mathcal{X}\), such as images, LiDAR data, or proprioceptive feedback \cite{2017_compare_actionspace_RL}, together with user commands \cite{2021_siekmann_sim2real_nonreference_perodicreward_DRL_e2e_LSTM_PPO_cassie} or pre-defined references \cite{2021_UCB_hybridrobotics_sim2real_referencebased_HZD_gaitlibrary_e2epolicy_drl_Cassie_lowpassfilter}, into joint-level control actions \(\mathcal{U}\). Here, \(\mathcal{X}\) represents the sensory input space, \(\mathcal{U}\) refers to the space of control actions, and \(\pi(\cdot)\) denotes the policy function. This framework obviates the need for manually decomposing the problem into sub-tasks, streamlining the control process.

End-to-end strategies primarily simplify the design of LL tracking to basic elements, such as a proportional–derivative (PD) controller. These methods can be broadly categorised according to their reliance on prior knowledge into two types: reference-based and reference-free. The locomotion skills developed through these diverse learning approaches exhibit considerable variation in performance and adaptability.

The following sections delve into various representation frameworks, exploring their characteristics, limitations, and strengths in comprehensive detail. To facilitate an understanding of these distinctions, Table \ref{tab:milestone_e2eframework} provides a succinct overview of the frameworks discussed.

\begin{table*}
 \begin{minipage}{\textwidth}
 \centering
   \caption{Summary and comparison of reference-based and reference-free learning approaches for the end-to-end framework. The dashed line in the implementation flow chart indicates optional steps.}\label{tab:milestone_e2eframework}
    \begin{tabular*}{\textheight}{@{\extracolsep{-9 pt}} *{9}{c} }
      \toprule
      \makecell{\textbf{Methods}} & \makecell{\textbf{Works}} &
      \makecell{\textbf{Capabilities}} &
      \makecell{\textbf{Characteristic}} & 
      \makecell{\textbf{Advantages and Disadvantages}} &
      \makecell{\textbf{Implementation Flow Chart}}  \\\midrule

        \makecell{ \begin{minipage}[t]{0.08\textwidth} \centering  Residual learning
        \end{minipage} }
         
        &\makecell{\cite{2018_xie_OSUDRL_feedback_DRL_reference-based}\\\cite{2020_Xie_firstsim2real_}}
        
        & \makecell{Forward walk\\unidirectional walk}
        
         & \makecell{ \begin{minipage}[t]{0.18\textwidth} Adding a residual term to the known motor positions at the current time step.
        \end{minipage} }
        
         & \makecell{ \begin{minipage}[t]{0.23\textwidth} {
         \textbf{A:} Fast convergence speed
         \vspace{1mm} 
         \hrule 
         \vspace{1mm} 
         \textbf{D:} Requires high-quality predefined reference, limits to specific motions, and lacks robustness to complex terrains.}
         \end{minipage}}
         
         & \makecell{ \includegraphics[width=0.26\linewidth, trim=9.8cm 8.8cm 10.8cm 9.8cm, clip]{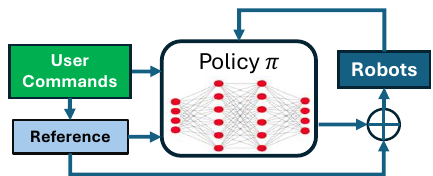} }\\
         \midrule
      \makecell{ \begin{minipage}[t]{0.08\textwidth} \centering Guided learning
        \end{minipage} }
         
        &\makecell{\cite{2020_Siekmann_drl_e2e_s2r_cassie}\\ \cite{2021_UCB_hybridrobotics_sim2real_referencebased_HZD_gaitlibrary_e2epolicy_drl_Cassie_lowpassfilter}\\\cite{2023_UCB_Cassie_RL_Jumping}\\\cite{2024_li_ucb_unifiedframework_PPO_IOput_teacherstudentcompare_IOhistory_e2etraining}}
        
        &\makecell{Forward walk\\Versatile walk\\Versatile jump\\Versatile motions}
        
         & \makecell{ \begin{minipage}[t]{0.18\textwidth} Mimic the predefined reference and directly specifies joint-level commands.
        \end{minipage} }
        
         & \makecell{ \begin{minipage}[t]{0.23\textwidth} {
         \textbf{A:} Accelerates the learning process and is robust across terrains.
         \vspace{1mm} 
         \hrule 
         \vspace{1mm} 
         \textbf{D:} Limits to the predefined motions and lacks adaptability to unforeseen changes in environment.}
         \end{minipage}}
         
         & \makecell{ \includegraphics[width=0.26\linewidth, trim=9.8cm 9cm 10.8cm 10cm, clip]{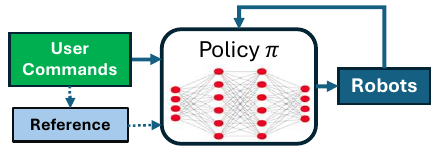} }
         \\
         \midrule

      \makecell{ \begin{minipage}[t]{0.1\textwidth} \centering Reference-free learning
        \end{minipage} }
         
        &\makecell{\cite{2021_siekmann_sim2real_nonreference_perodicreward_DRL_e2e_LSTM_PPO_cassie}\\\cite{2021_siekmann_blind_DRL_stair}\\\cite{2022_OSUDRL_steppingstone_referencefree_predictionfeasiblefootsteps_camera_benchmark}}
        &\makecell{Periodic motions\\Stepping stones walk\\Visual walk}
         & \makecell{ \begin{minipage}[t]{0.18\textwidth} Learn locomotion skills from scratch without any prior knowledge.
        \end{minipage} }
        &\makecell{ \begin{minipage}[t]{0.23\textwidth} {\textbf{A:} High potential for gait exploration, highly robust to complex terrain
         \vspace{1mm} 
         \hrule 
         \vspace{1mm} 
         \textbf{D:} Requires intensive reward shaping for gait patterns and is relatively expensive in computational resources}
         \end{minipage}}

         & \makecell{ \includegraphics[width=0.26\linewidth, trim=9.8cm 9cm 10.8cm 10cm, clip]{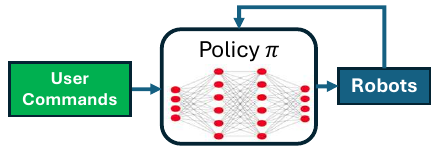} }
      \\
      \midrule
    \end{tabular*}

    \footnotetext{\textbf{Forward Walk} involves bipeds walking straight ahead. \textbf{Unidirectional Walk} enables bipeds to move forward and backward within a range of desired velocities. Omnidirectional Walk grants bipeds the ability to walk in any direction. \textbf{Versatile Walk} allows the bipeds to walk forward, backward, turn, and move sideways, providing extensive movement capabilities. \textbf{Periodic Motions} entails the execution of various repeated gait patterns, such as walking, hopping, or galloping. \textbf{Versatile Jump} refers to jumping towards different desired targets. \textbf{Versatile Motions} cover performing a broad array of motions, both periodic and aperiodic such as jumping.}
    \end{minipage}
\end{table*}

\subsection{Reference-based learning}
Reference-based learning leverages prior knowledge generated offline through methods such as TO or motion capture systems. This predefined reference typically includes data related to the robot's joint movements or pre-planned trajectories, serving as a foundation for the policy to develop locomotion skills by following these established motion patterns. Generally, this approach can be divided into two primary methods: (i) residual learning and (ii) guided learning.

\subsubsection{\textbf{Residual learning}} 

The proposed framework utilises a policy that modifies motor commands by applying action offsets based on the current reference joint positions, allowing the biped robot to achieve dynamic locomotion through error compensation. The state space includes proprioceptive information such as trunk position, orientation, velocity, angular velocity, joint angles, and joint velocities, providing the necessary sensory data for real-time adjustments. Actions are defined by offsets, $\delta a$, which represent deviations from the predefined desired joint positions, $\hat{a}$, with the final motor commands represented as $a = \hat{a} + \delta a$. The reward function encourages the policy to optimise locomotion performance by considering (a) how closely the robot’s active joint angles match the reference angles, (b) how effectively the robot responds to user commands, and (c) additional terms that further enhance the stability of the robot’s movements. This holistic approach enables the biped robot to adapt to various dynamic conditions while maintaining balance and control.

Introduced in 2018, a residual learning framework for the bipedal robot Cassie marked a significant advancement \cite{2018_xie_OSUDRL_feedback_DRL_reference-based}. This framework allowed the robot to walk forward by incorporating a policy trained via Proximal Policy Optimisation (PPO) algorithms, as detailed in Appendix \ref{Appendix:DRL_algorithm}. The policy receives the robot's states and reference inputs, outputting a residual term that augments the reference at the current timestep. These modified references are then processed by a PD controller to set the desired joint positions. 
Although this framework has improved the robot’s ability to perform tasks beyond standing \cite{2018_yang_push_recover_wholebody}, it has yet to be physically deployed on a bipedal robot. As a result, it remains impractical for managing walking at varying speeds and is constrained to movement in a single direction.

To transition this framework to a real robot, a sim-to-real strategy based on the previous model was demonstrated, where the policy, trained through a residual learning approach, was subsequently applied to a physical bipedal robot \cite{2020_Xie_firstsim2real_}. Compared to model-based methods, this training policy achieves faster running speeds on the same platform, underlining the considerable potential of DRL-based frameworks. However, the robot's movements remain constrained to merely walking forward or backward. 

A unique residual learning approach was introduced to enable omnidirectional walking, where the policy adds a residual term to the current joint positions, allowing gradual omnidirectional walking \cite{2021_Diego_deepwalk_curriculum_learning_sim2real}. In this case, the desired reference is the robot's current joint positions, which makes the approach distinctive. However, this also limits the policy’s ability to explore more diverse motions, restricting it to a single slow walking pattern.

Residual learning enhances an existing control policy by taking current joint positions or states and applying a residual action to adjust reference actions for better performance. Compared to other learning approaches that directly output joint positions, it is highly sample efficient \cite{2021_duan_DRL_task-spaceaction_hiarachycotnrolscheme_inversedynamiccontroller}. However, when predefined references are unstable or of low quality, residual learning may struggle, especially on complex terrains, as the action space is bounded by the reference, limiting the ability to handle unpredictable or uneven terrains.

\subsubsection{\textbf{Guided learning}} 
\label{subsec:guided_learning}
Guided learning trains policies to directly output the desired joint-level commands as actions $a$, without relying on the addition of a residual term. The state space is the same as the residual-learning approach. In this approach, the reward structure is centred on accurately imitating predefined reference trajectories, ensuring precise alignment between the policy output and the reference motion.

A sim-to-real framework that employs periodic references to initiate the training phase was proposed in \cite{2020_Siekmann_drl_e2e_s2r_cassie}. In this framework, the action space directly maps to the joint angles, and desired joint positions are managed by joint PD controllers. The framework also incorporates a Long Short-Term Memory (LSTM) network, as detailed in the Appendix \ref{Appendix:DRL_algorithm}, which is synchronised with periodic time inputs. However, this model is limited to a single locomotion goal: forward walking. A more diverse and robust walking DRL framework that includes a Hybrid Zero Dynamics (HZD) gait library was demonstrated \cite{2021_UCB_hybridrobotics_sim2real_referencebased_HZD_gaitlibrary_e2epolicy_drl_Cassie_lowpassfilter}, achieving a significant advancement by enabling a single end-to-end policy to facilitate walking, turning, and squatting.

Despite these advancements, the parameterisation of reference motions introduces constraints that limit the flexibility of the learning process and the policy's response to disturbances. To broaden the capabilities of guided learning policies, a framework capable of handling multiple targets, including jumping, was developed \cite{2023_UCB_Cassie_RL_Jumping}. This approach introduced a novel policy structure that integrates long-term input/output (I/O) encoding, complemented by a multi-stage training methodology that enables the execution of complex jumping manoeuvrers. An adversarial motion priors approach, employing a style reward mechanism, was also introduced to facilitate the acquisition of user-specified gait behaviours \cite{2024_Zhang_wholebody_adversarial_motion_priors}. This method improves the training of high-dimensional simulated agents by replacing complex hand-designed reward functions with more intuitive controls.

While previous works primarily focused on specific locomotion skills, a unified framework that accommodates both periodic and non-periodic motions was further developed \cite{2024_li_ucb_unifiedframework_PPO_IOput_teacherstudentcompare_IOhistory_e2etraining} based on the foundational work in \cite{2023_UCB_Cassie_RL_Jumping}. This framework enhances the learning process by incorporating a wide range of locomotion skills and introducing a dual I/O history approach, marking a significant breakthrough in creating a robust, versatile, and dynamic end-to-end framework. However, experimental results indicate that the precision of locomotion features, such as velocity tracking, remains suboptimal.

Guided learning methods expedite the learning process by leveraging expert knowledge and demonstrating the capacity to achieve versatile and robust locomotion skills. Through the comprehensive evaluation \cite{2024_li_ucb_unifiedframework_PPO_IOput_teacherstudentcompare_IOhistory_e2etraining}, it is demonstrated that guided learning employs references without complete dependence on them. Conversely, residual learning exhibits failures or severe deviations when predicated on references of inferior quality. This shortfall stems from the framework’s dependency on adhering closely to the provided references, which narrows its learning capabilities.

However, the benefits of reference-based learning come with inherent limitations. Reliance on predefined trajectories often confines the policy to specific gaits, restricting its capacity to explore a broader range of motion possibilities \cite{2021_UCB_hybridrobotics_sim2real_referencebased_HZD_gaitlibrary_e2epolicy_drl_Cassie_lowpassfilter,2023_vanmarum_visionDRL_studentteacher_irregularterrain_PPO_periodicrewardfunction}. Moreover, such methods exhibit reduced adaptability when confronted with novel environments or unforeseen perturbations. These limitations are further compounded by the difficulty of acquiring high-quality and task-relevant demonstrations.



Common sources of prior knowledge include TO \cite{2019_tianyu_ATRIAS_PPO_joint0level_trajectory_learnedhighlevelpolicy, 2021_Kevin_gaitlibrary_heirarchy_reducedorder_gaitlibrary_sim2real,2021_UCB_hybridrobotics_sim2real_referencebased_HZD_gaitlibrary_e2epolicy_drl_Cassie_lowpassfilter,2024_li_ucb_unifiedframework_PPO_IOput_teacherstudentcompare_IOhistory_e2etraining}, human motion capture \cite{2024_speratebody_expressive_allmotion}, teleoperation \cite{2023_seo_deep_loco_manipulation,2024_stanford_humanplus_hierarchical_humanoidshadowingimitation}, and scripted controllers \cite{2018_deepmimic_animation_PPO_deepmimic_desrieddirection_physics-based_animation_ATLAS_task}. While informative, these demonstrations often require adaptation due to embodiment mismatch or limited generalisability. Motion retargeting \cite{2019_iit_wholebody_retargeting,2024_tang_humanmimic_adversarial,2017_Ayusawa_motionretarget_optimization}, as one of promising direction, addresses this by converting human-centric motions into robot-feasible trajectories, yet it still struggles with preserving fidelity and adapting across morphologies. 

Ultimately, the success of guided learning relies not only on using references but on accessing high-quality, adaptable demonstrations that generalise across tasks and platforms—highlighting a key challenge in advancing robust policy learning.

\subsection{Reference-free learning}

In reference-free learning, the policy is trained using a carefully crafted reward function rather than relying on predefined trajectories. This approach allows the policy to explore a wider range of gait patterns and adapt to unforeseen terrains, thereby enhancing innovation and flexibility within the learning process. The action space and observation space in this approach are similar to the guided-learning method; however, the reward structure differs significantly from the reference-based method. Instead of focusing on imitating predefined motions, the reward emphasises learning efficient gait patterns by capturing the distinctive characteristics of bipedal locomotion \cite{2023_vanmarum_visionDRL_studentteacher_irregularterrain_PPO_periodicrewardfunction}.

The concept of reference-free learning was initially explored using simulated physics engines with somewhat unrealistic bipedal models. A pioneering framework, which focused on learning symmetric gaits from scratch without the use of motion capture data, was developed and validated within a simulated environment \cite{2018_wenhao_yu_DRL_withoutpredefine_symmetrygait_PPO_jointanglePD_}. This framework introduced a novel term into the loss function and utilised a curriculum learning strategy to effectively shape gait patterns. Another significant advancement was made in developing a learning method that enabled a robot to navigate stepping stones using curriculum learning, focusing on a physical robot model (Cassie), though this has yet to be validated outside of simulation \cite{2020_zhaoming_drl_steppingstones_PPOwithactorcritic_referencefree_simulation}.

Considering the practical implementation of this approach, significant efforts have been made to develop sim-to-real reference-free frameworks, and their potential has been further explored on physical robots.
A notable example of such a framework accommodates various periodic motions, including walking, hopping, and galloping \cite{2021_siekmann_sim2real_nonreference_perodicreward_DRL_e2e_LSTM_PPO_cassie}. This framework employs periodic rewards to facilitate initial training within simulations before successfully transitioning to a physical robot. It has been further refined to adapt to diverse terrains and scenarios. For instance, robust blind walking on stairs was demonstrated through terrain randomisation techniques in \cite{2021_siekmann_blind_DRL_stair}. Additionally, the integration of a vision system has enhanced the framework's ability to precisely determine foot locations \cite{2022_sim2real_footstepconstraint_OSUDRL_specifytouchdownlocation_actorcritic_PPO_LSTM_plannar_transisionmodel_model-predictiveplanning_CNN_predictnexttdlocation}, thus enabling the robot to effectively navigate stepping stones \cite{2022_OSUDRL_steppingstone_referencefree_predictionfeasiblefootsteps_camera_benchmark}. Subsequent developments include the incorporation of a vision system equipped with height maps, leading to an end-to-end framework that more effectively generalises terrain information \cite{2023_duan_OSUDRL_heightmap_visionbased_hybrid}.

This approach to learning enables the exploration of novel solutions and strategies that might not be achievable through mere imitation of existing behaviours. However, the absence of reference guidance can render the learning process costly, time-consuming, and potentially infeasible for certain tasks. Moreover, the success of this method hinges critically on the design of the reward function, which presents significant challenges in specifying tasks such as jumping.

\section{Hierarchy framework}\label{Sec:Hierarchy_framework}

\begin{figure*}
    \centering
    \includegraphics[trim={3.3cm 8cm 4cm 7cm},clip,width=\textwidth]{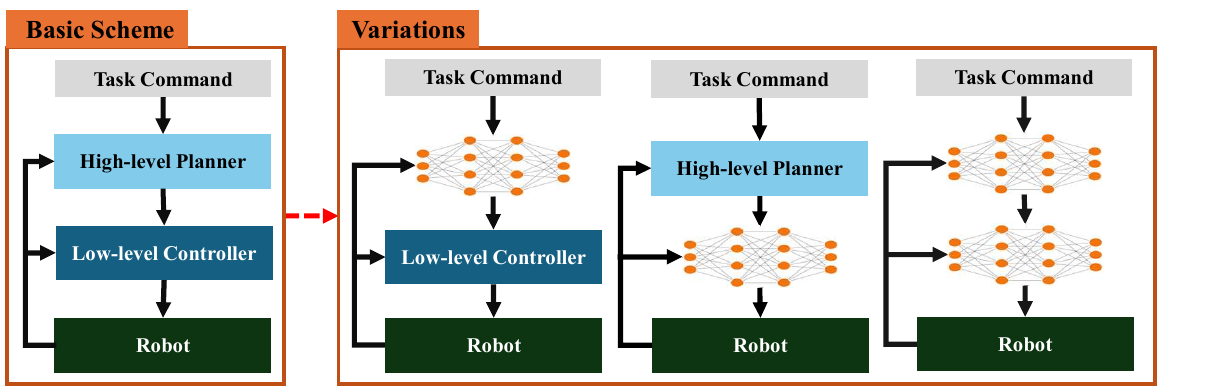}
    \caption{\textbf{Hierarchical control scheme diagram.} This figure illustrates a hierarchical control framework for a bipedal robot, comprising a basic scheme and three variations. (1) Basic scheme: The framework begins with a task command, followed by an HL planner and a LL controller, which ultimately drives the robot. Each module can be replaced with a learned policy, introducing adaptability across different control layers. (2) Variations (from left to right): (a) a deep planning hybrid scheme, in which the HL planner is learned; (b) a feedback DRL control hybrid scheme, with a learned LL controller; and (c) a learned hierarchical control scheme, where both layers are learned.}
    \label{fig:Hierarchy_scheme_flow_chart}
\end{figure*} 

Unlike end-to-end policies that directly map sensor inputs to motor outputs, hierarchical control schemes deconstruct locomotion challenges into discrete, manageable layers or stages of decision-making. Each layer within this structure is tasked with specific objectives, ranging from navigation to fundamental locomotion skills. This division not only enhances the framework's flexibility but also simplifies the problem-solving process for each policy. 

The architecture of a hierarchical framework typically comprises two principal modules: an HL planner and an LL controller. This modular approach allows for the substitution of each component with either a model-based method or a learning-based policy, further enhancing adaptability and customisation to specific needs. 

Communication between the layers in a hierarchical framework is achieved through the transmission of commands. The HL planner sets abstract goals, which the LL controller translates into specific actions, such as calculating joint movements to follow a desired trajectory. In return, the robot sends sensor data back to the HL planner, enabling real-time adjustments. The tasks handled by different layers often operate on varying time scales, adding complexity to synchronising communication between the layers.

Hierarchical frameworks can be classified into three distinct types based on the integration and function of their components:
\begin{enumerate}
    \item \textbf{Deep planning hybrid scheme:} This approach combines strategic, HL planning with dynamic LL execution, leveraging the strengths of both learning-based and traditional model-based methods.
    \item \textbf{Feedback DRL control hybrid scheme:} It focuses on integrating direct feedback control mechanisms with DRL, allowing for real-time adjustments and enhanced responsiveness.
    \item \textbf{Learned hierarchy scheme:} Entirely learning-driven, this scheme develops a layered decision-making hierarchy where each level is trained to optimise specific aspects of locomotion.
\end{enumerate}

These frameworks are illustrated in Fig. \ref{fig:Hierarchy_scheme_flow_chart}. Each type offers unique capabilities and exhibits distinct characteristics, albeit with limitations primarily due to the complexities involved in integrating diverse modules and their interactions.

For a concise overview, Table \ref{Tab: milestone_heirarchy framework} summarises the various frameworks, detailing their respective strengths, limitations, and primary characteristics. The subsequent sections will delve deeper into each of these frameworks, providing a thorough analysis of their operational mechanics and their application in real-world scenarios.

\begin{table*}
 \begin{minipage}{\textwidth}
 \centering
    \caption{Summary and comparison of hierarchical frameworks.}
    \label{Tab: milestone_heirarchy framework}
    \begin{tabular}{@{\extracolsep\fill} *{9}{c} }
      \toprule
      \makecell{\textbf{Control Scheme}} & 
      \makecell{\textbf{Works}} &
      \makecell{\textbf{Module}} &
      \makecell{\textbf{Characteristics}} &
      \makecell{\textbf{Advantages and Disadvantages}}\\\midrule
       \makecell{ \begin{minipage}[t]{0.1\textwidth} \centering Deep Planning Hybrid Scheme
        \end{minipage} }
         
        &\makecell{ \cite{2021_duan_DRL_task-spaceaction_hiarachycotnrolscheme_inversedynamiccontroller}\\ \cite{2023_template_taskspace_hierarchyscheme_reducedorderstateALIP_learnedhigherlevel_lowlevelinversedynamiccontroller}\\
        \cite{2024_gaspard_footstepnet_soccer}}

         & \makecell{Deep planning + ID \\ Deep planning + ID-QP \\ Deep planning + WPG }
        
         & \makecell{ \begin{minipage}[t]{0.18\textwidth} HL policy is learned to guide the LL controller to complete locomotion and navigation tasks.
        \end{minipage} }
        
         & \makecell{ \begin{minipage}[t]{0.26\textwidth} {\textbf{A: }Enhanced command tracking capabilities, generalised across different platforms, sampling-efficient, and robust.
         \vspace{1mm} 
         \hrule 
         \vspace{1mm} 
         \textbf{D: }Complicated system and communication between layers, requires a precise model, and lacks generalisation across different tasks.}
         \end{minipage}}
      \\\midrule
      \makecell{ \begin{minipage}[t]{0.15\textwidth} \centering Feedback DRL Control Hybrid Scheme
        \end{minipage} }
         
        &\makecell{\cite{2021_Kevin_gaitlibrary_heirarchy_reducedorder_gaitlibrary_sim2real}\\ \cite{2022_Singh_humanoid_e2erlwithfootplannar_hierarchyscheme_PPO_simulation_HRP-5P}\\ \cite{2023_wangsong_DRL_footsteptrakcing_hybridscheme_experiment_PPO_stair_task_LIPM_curriculumlearning}}

         & \makecell{Gait library + feedback policy\\Footstep planner + feedback policy\\Model-based planner + feedback policy}
        
         & \makecell{ \begin{minipage}[t]{0.18\textwidth} LL feedback policy receives non-learned HL planner as input to achieve locomotion skills. 
        \end{minipage} }
        
         & \makecell{ \begin{minipage}[t]{0.26\textwidth} {\textbf{A: }Short inference time, robust, navigation locomotion capabilities, interpretability.
         \vspace{1mm} 
         \hrule 
         \vspace{1mm} 
         \textbf{D: } Complicated system and communication between layers, which reduces sampling efficiency.}
         \end{minipage}}
    
       \\\midrule
      \makecell{ \begin{minipage}[t]{0.15\textwidth} \centering Learned Hierarchy Framework
\end{minipage} }
 
&\makecell{
    \cite{2017_Xuebinpeng_animation_deeploco_DRL_hiarachysystem_referencemotion_deepconvolutionalneuralnewtwork_jointangle_soccer.}\\ \cite{2023_Wei_learned_hierarchy_framework_wheeled_bipedalrobot}\\
    \cite{2024_stanford_humanplus_hierarchical_humanoidshadowingimitation}}

 & \makecell{HL policy + LL policy\\HL policy + LL policy\\HL policy + LL policy}

 & \makecell{ \begin{minipage}[t]{0.18\textwidth} Both HL planner and LL feedback controller are learned. LL policy focuses on basic locomotion skills; on the other side, HL policy learns navigation skills.
\end{minipage} }

 & \makecell{ \begin{minipage}[t]{0.26\textwidth} {\textbf{A: }  Provides layer flexibility, where each layer can be independently retrained and reused; alleviates the challenges associated with training an end-to-end policy.
 \vspace{1mm} 
 \hrule 
 \vspace{1mm} 
 \textbf{D: } Inefficient sim-to-real transfer, complicated interface between layers, and an expensive training process.}
 \end{minipage}}
      \\\midrule
    \end{tabular}

    \end{minipage}
\end{table*}

\subsection{Deep planning hybrid scheme}
In this scheme, robots are pre-equipped with the ability to execute basic locomotion skills such as walking, typically managed through model-based feedback controllers or interpretable methods. The addition of an HL learned layer focuses on strategic goals or the task space, enhancing locomotion capabilities and equipping the robot with advanced navigation abilities to effectively explore its environment.

Several studies have demonstrated the integration of an HL planner policy with a model-based controller to achieve tasks in world space. A notable framework optimises task space level performance, eschewing direct joint-level and balancing considerations \cite{2021_duan_DRL_task-spaceaction_hiarachycotnrolscheme_inversedynamiccontroller}. This system combines a residual learning planner with an inverse dynamics controller, enabling precise control over task-space commands to joint-level actions, thereby improving velocity tracking, foot touchdown location, and height control. Further advancements include a hybrid framework that merges HZD-based residual deep planning with model-based regulators to correct errors in learned trajectories, showcasing robustness, training efficiency, and effective velocity tracking \cite{2022_DRL_cascade_motionplanningpolicy_HZD_torso_PPOwithRNN}. These frameworks have been successfully transferred from simulation to reality and validated on robots such as Cassie.

However, the limitations imposed by residual learning constrained the agents' capacity to explore a broader array of possibilities. Building on previous work \cite{2022_DRL_cascade_motionplanningpolicy_HZD_torso_PPOwithRNN}, a more efficient hybrid framework was developed, which learns from scratch without reliance on prior knowledge \cite{2023_template_taskspace_hierarchyscheme_reducedorderstateALIP_learnedhigherlevel_lowlevelinversedynamiccontroller}. In this approach, a purely learning-based HL planner interacts with an LL controller using an Inverse Dynamics with Quadratic Programming formulation (ID-QP). This policy adeptly captures dynamic walking gaits through the use of reduced-order states and simplifies the learning trajectory. Demonstrating robustness and training efficiency, this framework has outperformed other models and was successfully generalised across various bipedal platforms, including Digit, Cassie, and RABBIT.

In parallel, several research teams have focused on developing navigation and locomotion planners for humanoid robots, leveraging onboard visual perception and learned control strategies. Recent work \cite{2024_gaspard_footstepnet_soccer} explored complex dynamic motion tasks such as playing soccer by integrating a learned policy with an online footstep planner that utilises weight positioning generation (WPG) to create a CoM trajectory. This configuration, coupled with a whole-body controller, enables dynamic activities like soccer shooting. Although these systems demonstrate promising coordination between perception, planning, and control, they remain limited in dynamic movement capability compared to full-sized humanoid robots, and thus primarily address navigation and task-level execution.

Regarding generalisation, these frameworks have shown potential for adaptation across different types of bipedal robots with minimal adjustments, demonstrating advanced user command tracking \cite{2023_template_taskspace_hierarchyscheme_reducedorderstateALIP_learnedhigherlevel_lowlevelinversedynamiccontroller} and sophisticated navigation capabilities \cite{2024_gaspard_footstepnet_soccer}. However, limitations are evident, notably the absence of capabilities for executing more complex and dynamic motions, such as jumping. Furthermore, while these systems adeptly navigate complex terrains with obstacles, footstep planning alone is insufficient without concurrent enhancements to the robot's overall locomotion capabilities. Moreover, the requisite communication between the two distinct layers of the hierarchical framework may introduce system complexities. Enhancing both navigation and dynamic locomotion capabilities within the HL planner remains a significant challenge.

\subsection{Feedback DRL control hybrid scheme}
In contrast to the comprehensive approach of end-to-end policies discussed in Section \ref{Sec:e2e_framework}, which excels in handling versatile locomotion skills and complex terrains with minimal inference time, the Feedback DRL Control Hybrid Scheme integrates DRL policies as LL controllers. These LL controllers, replacing traditional model-based feedback mechanisms, work in conjunction with HL planners that process terrain information, plan future walking paths, and maintain robust locomotion stability.

For instance, gait libraries, which provide predefined movement references based on user commands, have been integrated into such frameworks \cite{2021_Kevin_gaitlibrary_heirarchy_reducedorder_gaitlibrary_sim2real}. Despite the structured approach of using gait libraries, their static nature offers limited adaptability to changing terrains, diminishing their effectiveness. A more dynamic approach involves online planning, which has shown greater adaptability and efficiency. One notable framework combines a conventional foot planner with an LL DRL policy \cite{2022_Singh_humanoid_e2erlwithfootplannar_hierarchyscheme_PPO_simulation_HRP-5P}, delivering targeted footsteps and directional guidance to the robot, thereby enabling responsive and varied walking commands. Moreover, HL controllers can provide additional feedback to LL policies, incorporating CoM or end-feet information, either from model-based methods or other conventional control strategies. 
However, this work has not yet been transferred from simulation to real-world applications. Later, a similar structure featuring an HL foot planner and an LL DRL-based policy was proposed \cite{2023_wangsong_DRL_footsteptrakcing_hybridscheme_experiment_PPO_stair_task_LIPM_curriculumlearning}. This strategy not only achieved a successful sim-to-real transfer but also enabled the robot to navigate omnidirectionally and avoid obstacles.

A recent development has shown that focusing solely on foot placement might restrict the stability and adaptability of locomotion, particularly in complex maneuvers. A new framework integrates a model-based planner with a DRL feedback policy to enhance bipedal locomotion's agility and versatility, displaying improved performance \cite{2024_Li_model_based_LLDRL_hybrid_control_scheme}. This system employs a residual learning architecture, where the DRL policy's outputs are merged with the planner’s directives before being relayed to the PD controller. This integrated approach not only concerns itself with foot placement but also generates comprehensive trajectories for trunk position, orientation, and ankle yaw angle, enabling the robot to perform a wide array of locomotion skills including walking, squatting, turning, and stair climbing.

Compared to traditional model-based controllers, learned DRL policies provide a comprehensive closed-loop control strategy that does not rely on assumptions about terrain or robotic capabilities. These policies have demonstrated high efficiency in locomotion and accurate reference tracking \cite{quadRLLL}. Despite their extensive capabilities, such policies generally require short inference time, making DRL a preferred approach in scenarios where robustness is paramount or computational resources on the robot are limited. Nonetheless, these learning algorithms often face challenges in environments characterised by sparse rewards, where suitable footholds like gaps or stepping stones are infrequent \cite{quadRLLL}.

Additionally, an HL planner can process critical data such as terrain variations or obstacles and generate precise target locations for feet or desired walking paths, instead of detailed terrain data, which can significantly expedite the training process \cite{2023_wangsong_DRL_footsteptrakcing_hybridscheme_experiment_PPO_stair_task_LIPM_curriculumlearning}. This capability effectively addresses the navigational limitations observed in end-to-end frameworks. Moreover, unlike the deep planning hybrid scheme where modifications post-policy establishment can be cumbersome, this hybrid scheme offers enhanced flexibility for on-the-fly adjustments.

Despite the significant potential demonstrated by previous studies, integrating DRL-based controllers with sophisticated and complex HL planners still presents limitations compared to more integrated frameworks such as end-to-end and deep planning models. Specifically, complex HL model-based planners often require substantial computational resources to resolve problems, rely heavily on model assumptions, necessitate extensive training periods, demand large datasets for optimisation, and hinder rapid deployment and iterative enhancements \cite{quadRLLL}.

\subsection{Learned hierarchy framework}\label{Sec:learned_heirarchical_framework}
The Learned Hierarchy Framework merges a learned HL planner with an LL controller, focusing initially on refining LL policies to ensure balance and basic locomotion capabilities. Subsequently, an HL policy is developed to direct the robot towards specific targets, encapsulating a structured approach to robotic autonomy.

The genesis of this framework was within a physics engine, aimed at validating its efficiency through simulation \cite{2017_Xuebinpeng_animation_deeploco_DRL_hiarachysystem_referencemotion_deepconvolutionalneuralnewtwork_jointangle_soccer.}. In this setup, LL policies, informed by human motions or trajectories generated via TO, strive to track these trajectories as dictated by the HL planner while maintaining balance. An HL policy is then introduced, pre-trained with long-term task goals, to navigate the environment and identify optimal paths. This structure enabled sophisticated interactions such as guiding a biped to dribble a soccer ball towards a goal. The framework was later enhanced to include imitation learning (IL), facilitating the replication of dynamic human-like movements within the simulation environment \cite{2018_deepmimic_animation_PPO_deepmimic_desrieddirection_physics-based_animation_ATLAS_task}.

However, despite its structured and layered approach, which allows for the reuse of learned behaviours to achieve long-term objectives, these frameworks has been validated only in simulations. The interface designed manually between the HL planner and the LL controller sometimes leads to suboptimal behaviours, including stability issues like falling.

Expanding the application of this framework, a sim-to-real strategy for a wheeled bipedal robot was proposed, focusing the LL policy on balance and position tracking, while the HL policy enhances safety by aiding in collision avoidance and making strategic decisions based on the orientation of subgoals \cite{2023_Wei_learned_hierarchy_framework_wheeled_bipedalrobot}.


To fully leverage its potential, HumanPlus has been developed as a versatile framework for humanoid robots, integrating hierarchical learning, multimodal perception, and real-world imitation \cite{2024_stanford_humanplus_hierarchical_humanoidshadowingimitation}. It employs a two-layer structure, where HIT learns from human demonstrations, trained on AMASS, and HST acts as an LL tracking controller. Additionally, binocular RGB vision input enhances perception, enabling precise loco-manipulation and dynamic locomotion tasks such as jumping, walking, folding clothes, and rearranging objects. This shadowing-based IL approach improves adaptability, making it a promising framework for transferring human-like skills to robots.

Learning complex locomotion skills, particularly when incorporating navigation elements, presents a significant challenge in robotics. Decomposing these tasks into distinct locomotion and navigation components allows robots to tackle more intricate activities, such as dribbling a soccer ball \cite{2017_Xuebinpeng_animation_deeploco_DRL_hiarachysystem_referencemotion_deepconvolutionalneuralnewtwork_jointangle_soccer.}. As discussed in the previous section, the benefits of integrating RL-based planners with RL-based controllers have been effectively demonstrated. This combination enables the framework to adeptly manage a diverse array of environments and tasks.

Within such a framework, the HL policy is optimised for strategic planning and achieving specific goals. This optimisation allows for targeted enhancements depending on the tasks at hand. Moreover, the potential for continuous improvement and adaptation through further training ensures that the system can evolve over time, improving its efficiency and effectiveness in response to changing conditions or new objectives. Despite the theoretical advantages, the practical implementation of this type of sim-to-real application for bipedal robots remains largely unexplored.

Additionally, the training process for each policy within the hierarchy demands considerable computational resources \cite{2023_Wei_learned_hierarchy_framework_wheeled_bipedalrobot}. The intensive nature of this training can lead to a reliance on the simulation environment, potentially causing the system to overfit to specific scenarios and thereby fail to generalise to real-world conditions. This limitation highlights a significant hurdle that must be addressed to enhance the viability of learned hierarchy frameworks in practical applications.

Besides, for the general hierarchical framework, the transition from simulation to real-world scenarios is challenging, particularly due to the complexities involved in coordinating two layers within the control hierarchy. Ensuring seamless communication and cooperation between the HL planner and LL controller is essential to avoid operational discrepancies. The primary challenges include: (1) Task division complexity—while the HL planner handles strategy and provides abstract goals, the LL Controller manages precise execution, necessitating careful coordination to avoid functional overlap and conflicts. (2) Effective communication—the HL's abstract goals must be accurately interpreted and converted by the LL into real-time actions, especially in dynamic environments. (3) Task allocation—clear division of responsibilities between layers is crucial to prevent redundancy and ensure smooth system performance.

\section{Limitations and Challenges}
\label{Sec:limitations_challenges}

The end-to-end and hierarchical frameworks detailed in Sections~\ref{Sec:e2e_framework} and~\ref{Sec:Hierarchy_framework} represent the state of the art in DRL-based bipedal locomotion, demonstrating remarkable capabilities on specific tasks. However, a substantial gap remains between these task-oriented successes and the broader goal of achieving generalisation and adaptability across diverse morphologies, tasks, and environments. Bridging this gap requires more than incremental improvements—it demands the establishment of a unified framework that consolidates interfaces, training conventions, and evaluation protocols to systematically address the underlying limitations of current DRL pipelines.

As outlined in the following sections, the core challenges underlying this gap can be grouped into three interrelated aspects. At a foundational level, a primary difficulty involves the limitations and challenges in achieving both generalisation and precision (Section~\ref{subsec:generalisation_precision}). This is further complicated by the practical barrier of the sim-to-real gap in transferring policies from simulation to physical robots (Section~\ref{subsec:sim_to_real_gap}). Ultimately, these issues culminate in the critical challenges of ensuring safety and interpretability for robust deployment in real-world, safety-critical situations (Section~\ref{subsec:safety}).

\subsection{Generalisation and precision}
\label{subsec:generalisation_precision}


A central challenge in applying DRL to bipedal locomotion is the need to simultaneously achieve high generalisation across diverse skills and traverse to all kinds of terrains, and high precision in specific tasks. This remains a fundamental obstacle to realising truly unified and capable frameworks.

This capability gap is evident in the current literature. Many approaches excel at generalisation, demonstrating policies that enable versatile skills such as walking and jumping \cite{2021_UCB_hybridrobotics_sim2real_referencebased_HZD_gaitlibrary_e2epolicy_drl_Cassie_lowpassfilter, 2024_li_ucb_unifiedframework_PPO_IOput_teacherstudentcompare_IOhistory_e2etraining} and can transfer to different terrains \cite{2021_siekmann_blind_DRL_stair, 2022_OSUDRL_steppingstone_referencefree_predictionfeasiblefootsteps_camera_benchmark, 2023_radosavovic_realworld_transformer_NNmodel}. However, these generalised policies often lack the fidelity required for high-precision tasks such as exact foot placement \cite{2022_Singh_humanoid_e2erlwithfootplannar_hierarchyscheme_PPO_simulation_HRP-5P, 2022_OSUDRL_steppingstone_referencefree_predictionfeasiblefootsteps_camera_benchmark, 2022_sim2real_footstepconstraint_OSUDRL_specifytouchdownlocation_actorcritic_PPO_LSTM_plannar_transisionmodel_model-predictiveplanning_CNN_predictnexttdlocation} or maintaining a specific velocity with minimal error \cite{2023_vanmarum_visionDRL_studentteacher_irregularterrain_PPO_periodicrewardfunction}. Conversely, controllers specialised for narrow domains can achieve exceptional precision, as seen in jumping to a precise target \cite{2023_UCB_Cassie_RL_Jumping}, yet they cannot generalise these capabilities to a broader range of tasks. Thus, the development of a single unified framework that concurrently exhibits both broad competency and high fidelity remains largely unresolved.

This difficulty in uniting generalisation and precision is not arbitrary but stems from several key limitations inherent in current DRL paradigms, whether related to framework design, task formulation, or the training process itself:

\begin{itemize}
    \item \textbf{Limited terrain and gait patterns:} The failure to generalise is often a direct result of training on insufficiently diverse environments or with a restricted set of behaviours. Models trained on limited terrain are brittle when faced with novel surfaces, while a limited gait pattern library prevents adaptation to tasks requiring new motor skills.
    \item \textbf{Poor command tracking:} The learning signals for generalisation and precision are often in direct conflict. Generalisation requires permissive signals that allow the robot to adapt to varied terrains or recover from perturbations, whereas precision demands restrictive signals that minimise command-tracking error. Faced with these opposing objectives, a single policy is forced to compromise, which often leads to poor command tracking and the sacrifice of adaptability in favour of rigid, high-fidelity execution \cite{2023_Chongben_jumping_DDPG_multi-task_network_simulationonly,2024_li_ucb_unifiedframework_PPO_IOput_teacherstudentcompare_IOhistory_e2etraining}.
    \item \textbf{Inefficient sampling:} Underpinning the difficulty of solving both problems simultaneously is the inefficient sampling of most DRL algorithms \cite{2017_DRL_survey,2017_Schulman_PPO_instructure,2023_Omur_comparison_RLalgorithm_bipedallocomotion}. This problem is severely exacerbated in tasks that depend on sparse rewards, where feedback is infrequent and often only supports the success of the final task. Consequently, the immense amount of data required for an agent to explore, discover a successful strategy, and then refine it for both a diverse skill set for generalisation and the fine-grained control needed for precision is often computationally prohibitive, motivating massive parallel simulation merely to make training tractable \cite{2021_Sim_Isaac,2017_heess_emergence_animation_PPO_comprarisonrlalgorithm_differenttarrains,2017_Xuebinpeng_animation_deeploco_DRL_hiarachysystem_referencemotion_deepconvolutionalneuralnewtwork_jointangle_soccer.,2018_deepmimic_animation_PPO_deepmimic_desrieddirection_physics-based_animation_ATLAS_task}.
    \item \textbf{High-quality data scarcity:} As highlighted in Section~\ref{subsec:guided_learning}, the scarcity of high-quality demonstrations is a key bottleneck. Such data provide essential guidance for DRL, enabling policies to learn physically feasible and natural-looking gaits while avoiding unsafe exploration \cite{2018_deepmimic_animation_PPO_deepmimic_desrieddirection_physics-based_animation_ATLAS_task,2017_Xuebinpeng_animation_deeploco_DRL_hiarachysystem_referencemotion_deepconvolutionalneuralnewtwork_jointangle_soccer.,2020_motioncapture_referencebased_e2e_simulation_pybullet_augmentedrewarddesign}. This scarcity stems from the difficulty of transferring scalable human data due to embodiment mismatch \cite{2017_Ayusawa_motionretarget_optimization,2019_iit_wholebody_retargeting,2024_stanford_humanplus_hierarchical_humanoidshadowingimitation}, while generating feasible synthetic data via trajectory optimisation is often computationally expensive \cite{2016_LIPM_trajectoryoptimization,2019_tianyu_ATRIAS_PPO_joint0level_trajectory_learnedhighlevelpolicy}.
\end{itemize}

These fundamental limitations give rise to common algorithmic challenges, such as the need for complex reward engineering, and are directly reflected in the design of the field's dominant control architectures. End-to-end frameworks attempt a holistic solution, learning a single monolithic policy that must implicitly resolve all challenges simultaneously. While this approach can yield highly versatile and dynamic behaviours \cite{2023_radosavovic_realworld_transformer_NNmodel}, it directly confronts the immense difficulty of exploration from sparse rewards and the struggle of reconciling conflicting training objectives within unstable system dynamics. This often results in a lack of the fidelity and precision that hierarchical systems can enforce \cite{2024_li_ucb_unifiedframework_PPO_IOput_teacherstudentcompare_IOhistory_e2etraining}. Conversely, hierarchical frameworks are a direct architectural response to the lack of skill compositionality. By employing a ``divide and conquer'' strategy, they use an HL policy to sequence a library of LL, often model-based, controllers. This structure enforces precision and manages complex dynamics at a lower level \cite{2019_tianyu_ATRIAS_PPO_joint0level_trajectory_learnedhighlevelpolicy,2022_sim2real_footstepconstraint_OSUDRL_specifytouchdownlocation_actorcritic_PPO_LSTM_plannar_transisionmodel_model-predictiveplanning_CNN_predictnexttdlocation,2023_wangsong_DRL_footsteptrakcing_hybridscheme_experiment_PPO_stair_task_LIPM_curriculumlearning}. However, this results in a brittle system, imposing a strong prior that constrains the policy's freedom and limits its ability to generalise to situations not anticipated by the handcrafted controller \cite{2022_Singh_humanoid_e2erlwithfootplannar_hierarchyscheme_PPO_simulation_HRP-5P}.

\subsection{Challenges in transferring from simulation to reality}
\label{subsec:sim_to_real_gap}

Another challenge hindering the deployment of DRL policies on bipedal robots is the sim-to-real gap. This refers to the significant discrepancy between a policy's performance in a physics simulator and its performance on actual hardware. This gap is a critical obstacle because training directly on physical robots is often impractical. The millions of environmental interactions required for DRL would lead to accelerated mechanical wear, a risk of catastrophic failure, and require constant human supervision. While simulation offers a safe and efficient alternative, the ultimate goal of ``zero-shot'' transfer, where a policy works perfectly without any real-world fine-tuning, is rarely achieved.

A large body of research validates impressive locomotion skills purely within simulation, without attempting transfer to a physical system \cite{2017_Xuebinpeng_animation_deeploco_DRL_hiarachysystem_referencemotion_deepconvolutionalneuralnewtwork_jointangle_soccer.,2018_deepmimic_animation_PPO_deepmimic_desrieddirection_physics-based_animation_ATLAS_task,2021_DeepQ_DNQ_stepping_animationlonly_steppingstones,2017_Josh_drl_animation_imitationlearning_GRAIL_TRPO_MUJOCO,2023_Chongben_jumping_DDPG_multi-task_network_simulationonly}. Even when transfer is successful, it often comes with compromises. Many successful transfers are not truly ``zero-shot'' and rely on a subsequent phase of extensive real-world fine-tuning or manual parameter tuning \cite{2019_Yu_Phy_system_identification,2020_Xie_firstsim2real_}. In cases where policies do transfer without fine-tuning, they often exhibit a noticeable degradation in performance, where the robustness and agility seen in simulation are significantly lower in the real world \cite{2021_siekmann_blind_DRL_stair,2022_OSUDRL_steppingstone_referencefree_predictionfeasiblefootsteps_camera_benchmark,2020_park_recoverstability_drl_PPO_referencebased_simulation}.

This gap is caused by unavoidable differences between the virtual and physical worlds, which are especially problematic for dynamically unstable bipedal robots.

\begin{itemize}
    \item \textbf{Robot dynamics modelling and actuation:} Simulators struggle to replicate the complex dynamics of a physical bipedal robot, whose inherent instability makes it particularly sensitive to modelling errors. Factors such as motor friction, gear backlash, and precise link inertia are often simplified. 
    \item \textbf{Contact and terrain modelling:} Accurately simulating intermittent foot–ground contact is extremely difficult. A mismatch between simulated and real-world friction or surface properties can cause unexpected slips or bounces, leading to loss of balance.
    \item \textbf{Sensing and state estimation:} A simulated robot has access to perfect, noise-free state information. In the real world, these states must be estimated from noisy sensors such as IMUs and joint encoders \cite{2020_Xie_firstsim2real_,2019_Yu_Phy_system_identification}. For a bipedal robot, precise state estimation is critical for maintaining balance. 
\end{itemize}

Simulators such as Isaac Gym \cite{2021_Sim_Isaac}, RoboCup3D \cite{2003_Sim_Robocup}, OpenAI Gym \cite{2016_Sim_OpenAI}, and MuJoCo \cite{2012_Sim_Mujoco}, detailed in Appendix~\ref{Appendix:Training_Environment}, are widely used to train policies that closely mimic real-world physical conditions. These platforms use full-order dynamics to better represent the complex interactions robots face, and numerous sim-to-real frameworks \cite{2020_Xie_firstsim2real_,2022_Ashish_RMA_sim2real_UCB,2023_Rohan_sim2real_currentdecompose} have demonstrated efficient and high-performance results. Despite these advancements, a significant gap persists between simulation and reality, exacerbated by the approximations made in simulation and the unpredictability of physical environments.

\subsection{Safety-critical locomotion}
\label{subsec:safety}

Beyond performance metrics such as agility and robustness, the practical deployment of bipedal robots in human-centric environments is fundamentally contingent upon safety \cite{2024_tong_advancements_review,2021_Jenna_bipedrobot_review,2021_Justin_review_recentporgressinleggedrobot}. This includes ensuring the robot's own integrity to prevent costly damage, as well as guaranteeing the safety of the surrounding environment and any humans within it. While many existing frameworks have demonstrated impressive locomotion skills, they often prioritise performance over these safety considerations. This creates a critical barrier that separates success in controlled laboratory settings from reliable operation in the unpredictable real world.

\begin{itemize}
    \item \textbf{Blind locomotion policies:} Many current frameworks rely solely on internal sensors (proprioception) such as joint angles and IMU data \cite{2020_Siekmann_drl_e2e_s2r_cassie,2021_siekmann_blind_DRL_stair,2021_siekmann_sim2real_nonreference_perodicreward_DRL_e2e_LSTM_PPO_cassie}, creating a major safety risk. Lacking external perception, these robots cannot anticipate obstacles, slopes, or slippery surfaces, making them purely reactive and highly prone to failure. Despite these significant safety drawbacks, this approach is often adopted for several reasons: omitting vision simplifies the control problem to pure motor skills and avoids the computational cost of real-time visual processing. Moreover, since robust blind locomotion has already been demonstrated, vision is often treated as a supplementary component used to enhance task-specific precision \cite{2023_duan_OSUDRL_heightmap_visionbased_hybrid} or path planning \cite{2018_Tsunekawa_visualnavigation_hierarchy_DDPG-sim2real_NAO}, rather than a core requirement for basic stability.
    \item \textbf{Lack of physical constraint satisfaction:} Many DRL frameworks lack built-in mechanisms to guarantee physical constraint satisfaction. This gap has motivated constrained or safety-aware DRL that enforces limits via the learning objective or auxiliary safety modules—for example, Safe-RL on humanoids \cite{2020_Javier_safereinforcementlearning}, hybrid DRL with identified low-dimensional safety models \cite{2022_Li_UCB_safteystability_model-based_DRL_PPO}, footstep-constrained DRL policies \cite{2022_sim2real_footstepconstraint_OSUDRL_specifytouchdownlocation_actorcritic_PPO_LSTM_plannar_transisionmodel_model-predictiveplanning_CNN_predictnexttdlocation}, and reactive DRL steppers operating under feasibility constraints on uneven terrain \cite{2021_DeepQ_DNQ_stepping_animationlonly_steppingstones}. This limitation makes it difficult to prevent the robot from exceeding joint limits, applying excessive torques, or causing self-collisions, particularly when reacting to unexpected events. This is a key area where constrained RL could be applied.
\end{itemize}

In summary, the pursuit of performance in DRL has often sidelined critical safety issues. The prevalence of blind policies that cannot anticipate environmental hazards, combined with the lack of inherent mechanisms to enforce physical constraints, creates significant risk and hinders real-world deployment. While these challenges are considerable, they also define a clear path forward. The following section on Future Directions and Opportunities explores specific research avenues, such as vision-based learning and safe reinforcement learning, aimed at overcoming these safety barriers and enabling the development of truly robust and reliable bipedal robots.

\begin{figure*}[h]
    \centering
    \includegraphics[trim={0cm 0cm 0cm 0cm},clip,width=\textwidth]{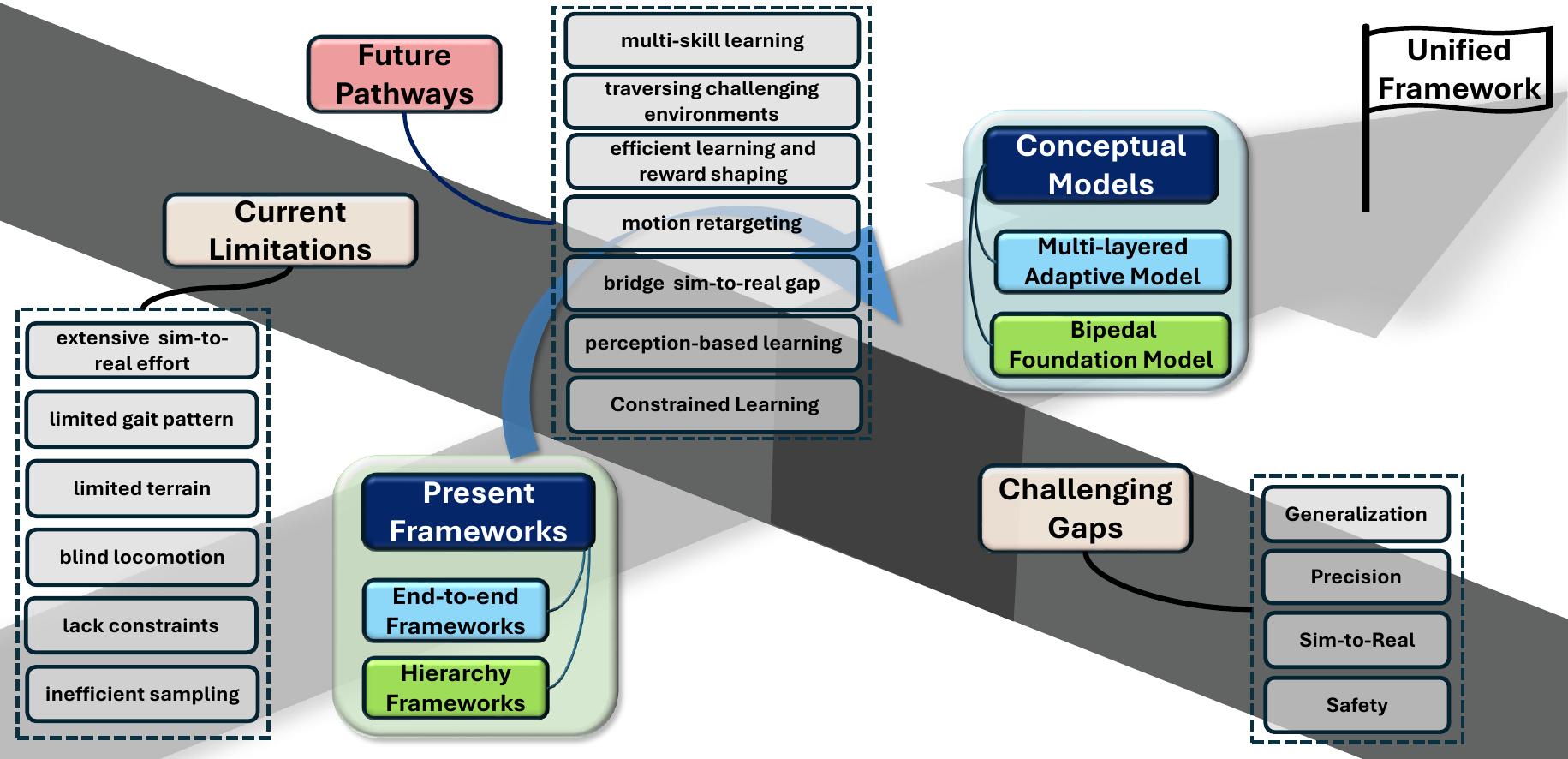}
       \caption{Towards a Unified Framework: This figure illustrates the logical progression from current DRL frameworks to future unified systems. It identifies the current limitations of existing end-to-end and hierarchical approaches, which motivate the exploration of specific Future Pathways. These pathways inform the design of two proposed conceptual models (i) Multi-Layered Adaptive Model (MLAM) and (ii) Bipedal Foundation Model (BFM) which represent potential blueprints for achieving a generalist, unified framework.}
    \label{fig:unified_framework_elements}
\end{figure*}

\section{Future Directions and Opportunities}
\label{Sec:future_directions_opportunities}

Following the analysis of the surveyed frameworks and their limitations, this section outlines a path forward for DRL-based bipedal locomotion by exploring both direct research avenues and emerging opportunities. We begin in Section~\ref{subsec:future_direction_bipedal_robot} by detailing research directions that directly respond to the challenges identified in Section~\ref{Sec:limitations_challenges}. Building on this foundation, Section~\ref{subsec:opportunites} broadens the scope to explore synergistic opportunities from related fields, such as loco-manipulation and the application of foundation models. These discussions culminate in Section~\ref{subsec:Unified_framework}, where we propose two conceptual models for a unified framework that represent the future evolution of the end-to-end and hierarchical paradigms.


\subsection{Pathways for bipedal locomotion}
\label{subsec:future_direction_bipedal_robot}

In relation to the research question introduced in Section~\ref{Sec:Introduction}, progress in DRL-based bipedal locomotion should be assessed not only through conventional metrics such as reward and success rate but also by broader system-level measures. These include generalisation breadth (across skills, terrains, and morphologies), precision in fidelity-critical tasks (e.g., command-tracking error and foot-placement accuracy), safety and constraint compliance (joint, torque, and contact feasibility), and efficiency or deployability (sample efficiency and on-robot inference latency).

These dimensions build directly upon the challenges outlined in Section~\ref{Sec:limitations_challenges} and together define the key pathways for advancing bipedal locomotion. The following subsections elaborate on these pathways, each addressing one or more of the above aspects to guide progress towards more generalisable and robust control systems.

\subsubsection{Multi-skill learning}
A fundamental goal for the next generation of bipedal robots is to move beyond the paradigm of single-task specialisation and towards versatile skill learning \cite{2023_radosavovic_realworld_transformer_NNmodel,2024_radosavovic_humanoid_locomotiontokenprediction_transformer,2024_li_ucb_unifiedframework_PPO_IOput_teacherstudentcompare_IOhistory_e2etraining}. This research direction focuses on enabling robots to acquire, adapt, and deploy a broad and varied repertoire of motor skills, allowing them to handle unforeseen situations and operate effectively in unstructured environments.

To achieve such versatility, researchers are pursuing several HL pathways, which can be broadly categorised into structured and holistic approaches. The structured approach focuses on explicit decomposition. A prominent example is hierarchical learning, where success depends on appropriately dividing responsibilities; for instance, an HL planner generates reference trajectories, while an LL DRL controller executes them robustly \cite{2024_stanford_humanplus_hierarchical_humanoidshadowingimitation,2023_template_taskspace_hierarchyscheme_reducedorderstateALIP_learnedhigherlevel_lowlevelinversedynamiccontroller}, as shown in Section~\ref{Sec:Hierarchy_framework}. Similarly, skill composition employs a supervisor policy to select and sequence LL experts to solve complex tasks \cite{2018_deepmimic_animation_PPO_deepmimic_desrieddirection_physics-based_animation_ATLAS_task}. A related technique, knowledge distillation, leverages experts by first training them and then distilling their capabilities into a single, compact generalist policy \cite{huang2024diffuseloco}.



\subsubsection{Traversing challenging environments}

The goal of versatile skill learning is to enable bipedal robots to traverse challenging, human-centric environments where their unique form offers an advantage. Validating capabilities on such terrains serves a crucial dual purpose. It tests a policy’s generalisation across diverse settings, including stairs and uneven ground, which is essential for real-world integration \cite{2021_siekmann_blind_DRL_stair,2023_vanmarum_visionDRL_studentteacher_irregularterrain_PPO_periodicrewardfunction,2023_radosavovic_realworld_transformer_NNmodel}. More critically, it benchmarks precision on treacherous paths such as stepping stones, which demand exact foot placement \cite{2022_OSUDRL_steppingstone_referencefree_predictionfeasiblefootsteps_camera_benchmark,2023_junheng_steppingstones_MPCWBC,2021_DeepQ_DNQ_stepping_animationlonly_steppingstones}. These environments are the ultimate test of both a robot’s skill repertoire and its control fidelity.


\subsubsection{Efficient learning and reward shaping} 

As detailed in Section~\ref{Sec:limitations_challenges}, while DRL has unlocked impressive capabilities in bipedal locomotion, its reliance on training from scratch leads to significant sample inefficiency \cite{2017_DRL_survey,2021_siekmann_sim2real_nonreference_perodicreward_DRL_e2e_LSTM_PPO_cassie}. Addressing this bottleneck is a crucial research frontier that calls for both more efficient algorithms and more robust reward designs.

To mitigate sample inefficiency for complex skills, several research pathways are being actively explored. A primary strategy is to leverage prior data rather than learning entirely from scratch. Leveraging prior knowledge provides strong guidance and reduces unsafe exploration by anchoring policies to feasible motion patterns \cite{2018_deepmimic_animation_PPO_deepmimic_desrieddirection_physics-based_animation_ATLAS_task,2018_xie_OSUDRL_feedback_DRL_reference-based,2024_stanford_humanplus_hierarchical_humanoidshadowingimitation}. Curriculum learning further organises training from simple to progressively harder tasks, for example standing and balancing before walking and running, which improves stability and convergence \cite{2020_zhaoming_drl_steppingstones_PPOwithactorcritic_referencefree_simulation,2021_Diego_deepwalk_curriculum_learning_sim2real,2023_wangsong_DRL_footsteptrakcing_hybridscheme_experiment_PPO_stair_task_LIPM_curriculumlearning}.

Complementing advances in algorithms is the design of effective and robust rewards. Manual reward engineering remains a significant obstacle, since small choices can induce reward hacking and lengthy tuning cycles \cite{2017_heess_emergence_animation_PPO_comprarisonrlalgorithm_differenttarrains,2018_deepmimic_animation_PPO_deepmimic_desrieddirection_physics-based_animation_ATLAS_task}. Phase-aware objectives are well established for cyclic gaits such as walking \cite{2021_siekmann_sim2real_nonreference_perodicreward_DRL_e2e_LSTM_PPO_cassie}, whereas reward design for non-periodic skills such as jumping is less standardised and often task specific \cite{2023_UCB_Cassie_RL_Jumping}. Promising directions reduce manual effort by adding higher-level guidance, including event-based terms, goal-conditioned objectives, and kinematic reference tracking \cite{2022_sim2real_footstepconstraint_OSUDRL_specifytouchdownlocation_actorcritic_PPO_LSTM_plannar_transisionmodel_model-predictiveplanning_CNN_predictnexttdlocation,2021_michael_velocity-basedcontrol_motioncapturedata,2018_yang_push_recover_wholebody}. Alternatively, learning rewards from data through inverse methods and related approaches aims to replace hand-crafted objectives with implicit ones inferred from demonstrations \cite{2016_GAIL_imitation_learning}. Together, these directions seek to minimise skill-specific tuning and improve the transferability and reliability of learned locomotion policies.

\subsubsection{Motion retargeting}
\label{subsec:motion_retarget}
As human-like agents, bipedal robots—especially humanoids—have the unique advantage of a morphology that is similar to our own. This presents a significant opportunity: the potential to learn from vast libraries of human motion data. While large-scale datasets such as AMASS \cite{2019_Naureen_AMASS} and Motion-X \cite{2024_motion_X_datasets} provide a wealth of such data, they are inherently human-centric and cannot be used directly, requiring substantial retargeting effort \cite{2024_speratebody_expressive_allmotion}. Therefore, motion retargeting emerges as a critical component to bridge this gap. The challenge of this pathway is not merely to transfer human movements to the robot, but to generate trajectories that are both high in stylistic fidelity and physically feasible, adhering to the robot's unique dynamics and constraints. Successfully developing these retargeting methods provides a scalable solution for accessing the data needed to train the natural and versatile generalist policies of the future.



\subsubsection{Bridging the gap from simulation to reality}

Strategies to bridge the sim-to-real gap generally follow two main philosophies. The first aims to train policies robust enough to tolerate the inevitable mismatch between simulation and reality, while the second focuses on minimising the gap itself by making the simulator a more faithful replica of the physical world.

The first approach seeks to reduce the discrepancy by improving the simulation's fidelity. This is often achieved through system identification (SI), where real-world robot data are used to fine-tune the simulator's parameters to create a more accurate “digital twin” \cite{2019_Yu_Phy_system_identification,2023_masuda_simtoreal_identification}. This can include explicitly learning complex actuator dynamics to model the motors’ behaviour \cite{2019_Yu_Phy_system_identification,2019_simtoreal_actuatornet_quadrupedrobot}. Other methods, such as designing specialised feedback controllers \cite{2021_Casillo_cyberbotics_hierarchyscheme_bezierpolynominals}, also contribute by making the system less sensitive to residual modelling errors.

In contrast, the second philosophy accepts that simulations will always be imperfect and instead focuses on creating highly adaptive, robust policies. The primary method here is DR, which forces a policy to generalize by training it across a wide range of simulated physical variations. Other various ways, such as through end-to-end training that uses measurement histories to adapt online like in RMA \cite{2022_Ashish_RMA_sim2real_UCB}, or via policy distillation, where a privileged ``teacher'' guides a ``student'' policy \cite{2023_vanmarum_visionDRL_studentteacher_irregularterrain_PPO_periodicrewardfunction} to have a knowledge of unknown information like friction . Additionally, techniques like adversarial motion priors \cite{2024_Zhang_wholebody_adversarial_motion_priors,2024_tang_humanmimic_adversarial} are used to ensure the learned behaviours are not just robust but also physically plausible.

Looking ahead, the ultimate goal remains achieving reliable zero-shot transfer, where no real-world fine-tuning is needed. Progress will depend on the co-development of higher-fidelity simulations, improved hardware, and more robust control policies inherently capable of handling real-world unpredictability. The synergy of these advancements will be crucial in finally closing the sim-to-real gap.

\subsubsection{Perception-conditioned locomotion}

Integrating exteroceptive sensors such as cameras and LiDAR enables bipedal robots to proactively plan footsteps, avoid obstacles, and adapt to upcoming terrain. This shift from reactive to anticipatory control is essential for navigating unstructured real-world environments.

The vision-based pathway is a human-inspired approach using RGB and depth cameras to capture rich data on colour, texture, and object appearance \cite{2022_OSUDRL_steppingstone_referencefree_predictionfeasiblefootsteps_camera_benchmark,2023_duan_OSUDRL_heightmap_visionbased_hybrid,2023_wangsong_DRL_footsteptrakcing_hybridscheme_experiment_PPO_stair_task_LIPM_curriculumlearning}. In contrast, LiDAR is an active sensing method that generates precise 3D point clouds of the terrain. While vision provides richer data but is sensitive to lighting, LiDAR offers robust geometric measurements without visual detail.

Based on this sensory data, current research is exploring two primary pathways for processing perceptual information for control. The first involves creating an intermediate geometric representation, such as a height map from scanners \cite{2023_duan_OSUDRL_heightmap_visionbased_hybrid}. This provides the policy with structured topographical data for effective footstep planning. The second is a more end-to-end approach, which utilises direct vision inputs such as RGB or depth images as inputs to the RL policy for real-time decision-making \cite{2018_Tsunekawa_visualnavigation_hierarchy_DDPG-sim2real_NAO,2022_byravan_nerf2real_basketball}. The former offers interpretability, while the latter promises more nuanced, reactive behaviours learned directly from raw perception.

Future progress requires advancing both pathways: building richer, semantic world representations and improving the efficiency of direct perception-to-action policies. Solving the underlying challenges of real-time processing and the perceptual sim-to-real gap will be crucial for enabling truly adaptive locomotion in complex, real-world scenarios.


\subsubsection{Constrained learning}
While the previously discussed pathways focus on enhancing a robot’s capabilities, a parallel and equally critical frontier is ensuring that these capabilities are exercised safely and reliably. To formally integrate safety, modern approaches can be grouped by how they handle constraints: soft constraints that guide the policy through costs and hard constraints that strictly limit actions \cite{2020_Javier_safereinforcementlearning,2022_Li_UCB_safteystability_model-based_DRL_PPO}.

Soft constraints encourage desirable behaviour and penalise undesirable behaviour without forbidding it. They are well suited to preferences or efficiency goals, for example minimising energy use, limiting peak torques, or promoting smooth motion \cite{2020_Javier_safereinforcementlearning}. Hard constraints are inviolable rules that prevent catastrophic failures. They are essential for enforcing physical limits and protecting the robot and its environment, for example footstep feasibility, contact timing, joint and torque bounds, and collision avoidance \cite{2022_sim2real_footstepconstraint_OSUDRL_specifytouchdownlocation_actorcritic_PPO_LSTM_plannar_transisionmodel_model-predictiveplanning_CNN_predictnexttdlocation,2021_Casillo_cyberbotics_hierarchyscheme_bezierpolynominals}. A practical way to enforce hard constraints is to use safety filters or shields grounded in control theory, such as control barrier functions and related template model checks \cite{2016_Quan_gaitlibrary}.

In practice, a robust and trustworthy bipedal robot will likely combine both ideas. Soft constraints help a policy learn efficient and natural gaits, while hard constraints guarantee that it will not take catastrophic actions. This combination supports the transition from systems that are merely capable in laboratory settings to agents that are reliable, predictable, and safe for real-world deployment.

\subsection{Opportunities}
\label{subsec:opportunites}

\subsubsection{Leveraging foundation models for locomotion learning}

The recent rise of Foundation Models (FMs), such as Large Language Models (LLMs) and Vision–Language Models (VLMs), presents a transformative opportunity for bipedal locomotion. Their powerful reasoning capabilities are unlocking new approaches that go beyond traditional control methods, primarily by enabling sophisticated HL task planning and by providing novel solutions to shape the learning process itself, particularly in automated reward design.

As HL planners, FMs provide a reasoning engine that can bridge the gap between abstract human goals and LL motor execution. They can interpret complex linguistic commands or visual scenes and decompose them into a sequence of simpler, actionable commands for an LL policy to follow. This has been demonstrated effectively in legged robotics, where VLMs process raw sensory data to pass structured commands to motor controllers \cite{2024_VLM_PC}, creating a seamless link between strategic planning and physical action.

Furthermore, FMs create a significant opportunity to overcome one of the most persistent bottlenecks in DRL: reward design. Instead of tedious manual tuning, LLMs can dynamically generate or refine reward functions based on linguistic descriptions of task success. Research has shown that LLMs can translate human feedback into reward adjustments \cite{2023_kumar_wordscontrol_LLM} or even autonomously adjust rewards and control strategies to self-optimise for diverse terrains \cite{2024_LLM_AnyBipe}, drastically reducing human intervention.

The foremost opportunity lies in the deeper synergy between these roles. The integration of the HL symbolic reasoning of FMs with the LL, real-time control of DRL could create a new class of highly adaptive and flexible robots. As this rapidly evolving field progresses, as reviewed in \cite{2023_Fms_survey_robotics_Roya_applicationsChallengeandFuture}, we may see a paradigm shift towards more autonomous, self-learning humanoid robots that can understand, reason about, and adapt to the world with minimal human intervention.

 \subsubsection{Loco-manipulation tasks}
While achieving stable locomotion is a foundational challenge, a bipedal robot with only a lower body has limited practical utility, as it cannot physically interact with its environment. The evolution of modern humanoids to include complex upper bodies is a critical advancement that has unlocked the opportunity for loco-manipulation—the dynamic integration of movement and object interaction. Achieving such full-body coordination is now a key benchmark for creating truly adaptable systems, with tasks ranging from climbing and using tools to carrying objects while navigating, as highlighted by initiatives like the DARPA Robotics Challenge \cite{2015_DAPRA_Challenge_1}.

However, realising this opportunity is a significant challenge. Early studies, such as a ‘box transportation’ framework \cite{2023_dao_OSUDRL_loco-manipulation_hybrid_simtoreal}, often rely on inefficient, multi-policy solutions that lack visual perception. Furthermore, dynamically interacting with mobile objects like scooters or balls introduces even greater complexity \cite{2023_baltes_control_scooter, 2023_deepmind_haarnoja_playsoccer}.

These difficulties create significant research opportunities. One such opportunity lies in exploring hierarchical control approaches. By decomposing tasks into multiple layers, this method allows for precise, modular control over different components, which can enhance stability and adaptability to environmental variations \cite{2023_template_taskspace_hierarchyscheme_reducedorderstateALIP_learnedhigherlevel_lowlevelinversedynamiccontroller}.

Alternatively, a further research opportunity is the development of end-to-end learning frameworks, which offer a more scalable solution. Using techniques like curriculum learning and imitation from human motion-capture data \cite{2021_Diego_deepwalk_curriculum_learning_sim2real,2023_wangsong_DRL_footsteptrakcing_hybridscheme_experiment_PPO_stair_task_LIPM_curriculumlearning,2024_li_ucb_unifiedframework_PPO_IOput_teacherstudentcompare_IOhistory_e2etraining,2023_seo_deep_loco_manipulation,2024_Zhang_wholebody_adversarial_motion_priors,2024_speratebody_expressive_allmotion}, a single, unified policy can be trained to handle diverse loco-manipulation tasks, representing a promising avenue of research for creating truly versatile agents.

 \subsubsection{Insights from quadruped robots}
While DRL remains an emerging technology in bipedal robotics, it has firmly established its presence in the realm of quadruped robots, another category of legged systems. The diversity of frameworks developed for quadrupeds ranges from end-to-end, model-based RL designed for training in real-world scenarios, where unpredictable dynamics often prevail \cite{quadModelBased1, quadModelBased2}, to systems that include the modelling of deformable terrain to enhance locomotion over compliant surfaces \cite{quadDeformable}. Furthermore, dynamic quadruped models facilitate highly adaptable policies \cite{quadModular, humphreys2024bio}, and sophisticated acrobatic motions are achieved through IL \cite{quadImitation2}.

The domain of quadruped DRL has also seen significant advancements in complex hierarchical frameworks that integrate vision-based systems. To date, two primary versions of such hierarchical frameworks have been developed: one where a deep-planning module is paired with model-based control \cite{quadRLPlanning} within a deep-planning hybrid scheme, and another that combines model-based planning with LL DRL control \cite{quadRLLL, quadmpcref} within a feedback DRL control hybrid scheme. The latter has shown substantial efficacy; it employs an MPC to generate reference motions, which are then followed by an LL feedback DRL policy. Additionally, the Terrain-aware Motion Generation for Legged Robots module \cite{TAMOLS} enhances the MPC and DRL policy by providing terrain height maps for effective foothold placements across diverse environments, including those not encountered during training. However, similar hierarchical hybrid control schemes have not been thoroughly investigated within the field of bipedal locomotion.

Quadruped DRL frameworks are predominantly designed to navigate complex terrains, but efforts to extend their capabilities to other tasks are under way. These include mimicking real animals through motion-capture data and IL \cite{quadAnimal1, quadAnimal2}, as well as augmenting quadrupeds with manipulation abilities. This is achieved either by adding a manipulator \cite{quadManip1, quadManip2} or by using the robots’ legs \cite{quadManip3}. Notably, the research presented in \cite{quadManip2} demonstrates that loco-manipulation tasks can be effectively managed using a single, unified, end-to-end framework.

Despite the progress in quadruped DRL, similar advancements have been limited for bipedal robots, particularly in loco-manipulation tasks and vision-based DRL frameworks; a combination of their inherent instability, lack of accessibility to researchers, and high mechanical complexity can be attributed to this disparity between quadruped and bipedal robots. Establishing a unified framework could bridge this gap—an essential step, given the integral role of bipedal robots with upper bodies in developing fully functional humanoid systems. Moreover, the potential of hybrid frameworks that combine model-based and DRL-based methods in bipedal robots remains largely untapped.

\subsection{Conceptual models for unified frameworks} \label{subsec:Unified_framework}

Motivated by our survey and the current state of the art, we propose two conceptual models, intended as reference designs, towards a unified locomotion framework. They build on end-to-end and hierarchical paradigms and offer complementary routes to scalable, generalisable architectures, rather than fully realised systems.

\begin{itemize}
\item \textbf{Bipedal Foundation Models (BFMs):} large-scale, pre-trained models that map perception directly to action through representation learning. Trained on diverse data across tasks and embodiments, BFMs aim to enable generalist locomotion control by supporting rapid adaptation via fine-tuning.
\item \textbf{Multi-Layer Adaptive Models (MLAMs):} modular, hierarchical architectures that span from HL planning to LL control, with each layer producing interpretable intermediate outputs. MLAMs are designed to integrate, substitute, and coordinate diverse policies, enabling flexible and adaptive responses across tasks and embodiments.
\end{itemize}

In the following sections, we will analyse each of these conceptual models in detail, evaluating their respective strengths and challenges in the pursuit of a unified framework.

\subsubsection{Bipedal foundation models}

Inspired by Robot Foundation Models (RFMs)~\cite{2023_Fms_survey_robotics_Roya_applicationsChallengeandFuture, 2023_generalpurposerobot_FMs}, we propose the concept of BFMs as large-scale, general-purpose models tailored for bipedal locomotion. A BFM would be a large-scale model pre-trained specifically to learn the shared motion priors of dynamic balance and movement across a vast range of bipedal tasks and physical embodiments. Unlike traditional policies trained from scratch, a BFM would provide a foundational understanding of stable locomotion, directly tackling the core difficulties that make bipeds distinct from other robots. Architecturally, we envision such a model comprising a multi-modal embedding module, a shared backbone like a transformer, and an action decoder, drawing inspiration from models like RT-2~\cite{2023_VLA_Rt-2}.

The proposed BFM paradigm would operate in two stages. First, IL on diverse datasets would establish the generalisable foundation. Second, DRL would be repurposed as an efficient fine-tuning mechanism to adapt these general priors to the specific, and often unforgiving, dynamics of a physical robot. The potential of this approach is highlighted by recent works, with frameworks like FLaRe~\cite{2024_FLaRe_RLfinetune_foundationmodel} enhancing generalisation for long-horizon tasks, MOTO~\cite{2024_MOTO_RLfinetune_foundationmodel} enabling effective offline-to-online adaptation from images, and AdA~\cite{2023_RL_finetune_RMFs} demonstrating in-context adaptation to novel environments. Collectively, these approaches underscore DRL not only as a simple tuning tool but as a central mechanism for grounding abstract foundation model priors into executable, platform-specific control policies.

However, realising the BFM concept for bipeds presents significant challenges. The DRL fine-tuning stage can be costly and risky on physical hardware, and policies may overfit to narrow dynamics or catastrophically forget the generalisable priors acquired during pre-training~\cite{2024_FLaRe_RLfinetune_foundationmodel, 2023_RL_finetune_RMFs}. Furthermore, as detailed in Sections~\ref{subsec:generalisation_precision} and~\ref{subsec:motion_retarget}, the scarcity of high-quality, large-scale data remains a fundamental bottleneck, as most existing datasets are human-centric and require significant adaptation before they can be used.

\subsubsection{Multi-layered adaptive models}


As a complementary path to BFMs, we propose the concept of MLAMs. Rather than relying on large-scale pre-training, this conceptual framework would adopt a modular, hierarchical approach. The idea is to extend conventional hierarchical frameworks (discussed in Section \ref{Sec:Hierarchy_framework}) with explicitly adaptive layers, allowing for the dynamic composition of specialised policies. The core principle of this concept would be modularity, enabling each layer to be independently optimised or replaced and providing interpretable outputs at each stage.

A key feature we envision for MLAMs is their capacity to dynamically compose adaptive modules for each control tier. Each layer processes context-specific inputs and outputs interpretable commands. The HL reasoning layer leverages large pre-trained models such as LLMs and VLMs \cite{2022_DoAsICan,2022_codeaspolicy_google} to parse commands into sub-tasks. For instance, Vision–Language Model Predictive Control \cite{2024_VLM_PC} has been effective in quadrupedal robots, integrating linguistic and visual inputs to optimise HL task planning. By leveraging LLMs, a unified framework could seamlessly bridge HL strategic planning with detailed task execution.

The mid-level planner selects or synthesises motions via learned motion libraries \cite{2021_Kevin_gaitlibrary_heirarchy_reducedorder_gaitlibrary_sim2real,2021_UCB_hybridrobotics_sim2real_referencebased_HZD_gaitlibrary_e2epolicy_drl_Cassie_lowpassfilter} or DRL-based planners \cite{2021_Mohammadreza_DRL_improveparameters_forplanner}. The LL control layer comprises various modular controllers, dynamically selected and composed based on task-specific demands. These include locomotion primitives like walking and climbing \cite{2024_yangOu_LLM_hierarchical_quadrupedrobot}, adaptive tracking controllers for whole-body tracking \cite{2024_stanford_humanplus_hierarchical_humanoidshadowingimitation}, and imitation-based skills such as kicking and dancing \cite{2018_deepmimic_animation_PPO_deepmimic_desrieddirection_physics-based_animation_ATLAS_task}, by utilising RL, IL, or model-based methods. This layered architecture is exemplified by recent work on quadrupedal robots, where LLMs are used to translate HL commands into robust and flexible real-world behaviours \cite{2024_yangOu_LLM_hierarchical_quadrupedrobot}.

However, realising the MLAM concept would introduce challenges distinct from BFMs. Such a framework would depend heavily on real-time multi-modal perception, which complicates data alignment across layers with differing timescales and abstraction levels~\cite{TAMOLS}. Additionally, the computational latency incurred by HL reasoning modules like LLMs~\cite{2024_yangOu_LLM_hierarchical_quadrupedrobot} would pose limitations for tasks needing rapid reactions.

\section{Conclusion}\label{sec:conclusion}

Despite significant progress in DRL for robotics, a substantial gap remains between current achievements and the development of a unified framework capable of efficiently handling a wide range of complex tasks. DRL research is generally divided into two main control schemes: end-to-end and hierarchical frameworks. End-to-end frameworks have demonstrated success in handling diverse locomotion skills \cite{2024_li_ucb_unifiedframework_PPO_IOput_teacherstudentcompare_IOhistory_e2etraining}, climbing stairs \cite{2021_siekmann_blind_DRL_stair}, and navigating challenging terrains such as stepping stones \cite{2022_OSUDRL_steppingstone_referencefree_predictionfeasiblefootsteps_camera_benchmark}. Meanwhile, hierarchical frameworks provide enhanced capabilities, particularly in managing both locomotion and navigation tasks simultaneously.

Each framework contributes unique strengths to the pursuit of a unified framework. End-to-end approaches simplify control by directly mapping inputs to outputs, while reference-based and reference-free learning methods provide the versatility required for robots to acquire diverse locomotion skills. In contrast, hierarchical frameworks improve flexibility by structuring control into layers, allowing modular task decomposition and hybrid strategies.

While DRL has enabled remarkable progress, our survey concludes that current frameworks face key limitations, including the tension between multi-skill generalisation and task-specific precision, the persistent sim-to-real gap, and critical safety concerns. To address these challenges, this survey synthesises specific pathways for future research and identifies key opportunities for cross-pollination from related fields, such as FMs, loco-manipulation, and quadrupedal robotics. These insights culminate in our proposal of two conceptual frameworks: the BFMs, extending the end-to-end paradigm, and the MLAMs, evolving from the hierarchical approach, which together offer distinct blueprints for the next generation of generalist bipedal controllers.

\begin{appendices}

\section{Deep reinforcement learning algorithms}\label{Appendix:DRL_algorithm}

\begin{figure}[b]
    \centering
    \includegraphics[trim={2cm 4.1cm 0cm 4.2cm},clip,,width=\linewidth]{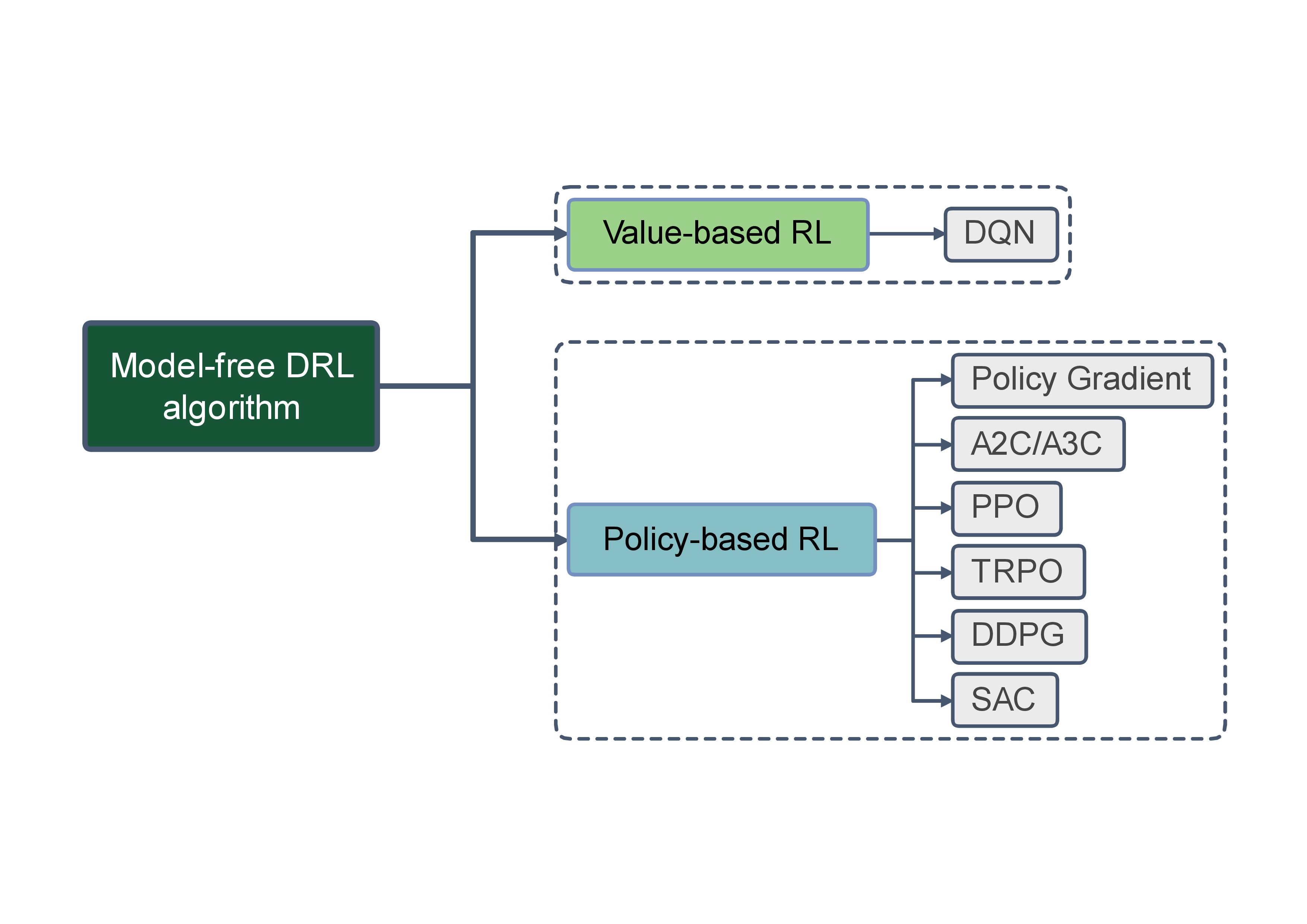}
    \caption{\textbf{Diagram for RL algorithms catalogue} } 
    \label{fig:catalogue of RL}
\end{figure} 

The advancement and development of RL are crucial for bipedal locomotion. Specifically, advances in deep learning provide deep NNs that serve as function approximators, enabling RL to handle tasks characterised by high-dimensional and continuous spaces by efficiently discovering condensed, low-dimensional representations of complex data. In comparison with other robots of different morphologies, such as wheeled robots, bipedal robots possess far higher DoFs and continuously interact with their environments, which results in greater demands on DRL algorithms. In particular, within legged locomotion, policy-gradient-based algorithms are prevalent in bipedal locomotion research.

Designing an effective NN architecture is essential for tackling complex bipedal locomotion tasks. Multi-Layer Perceptrons (MLPs), a fundamental NN structure, excel in straightforward regression tasks with lower computational resource requirements. A comprehensive comparison between MLPs and the memory-based NN LSTM reveals that MLPs have an advantage in convergence speed for many tasks \cite{2023_Rohan_sim2real_currentdecompose}. However, LSTMs, as variants of Recurrent Neural Networks (RNNs), are adept at processing time-associated data, effectively relating different states across time and modelling key physical properties vital for periodic gaits \cite{2021_siekmann_sim2real_nonreference_perodicreward_DRL_e2e_LSTM_PPO_cassie} and successful sim-to-real transfer in bipedal locomotion. Additionally, Convolutional Neural Networks (CNNs) specialise in spatial data processing, particularly for image-related tasks, making them highly suitable for environments where visual perception is crucial. This diversity of NN architectures highlights the importance of selecting an appropriate model based on the specific requirements of bipedal locomotion tasks.

Considering DRL algorithms, recent bipedal locomotion studies have focused on model-free RL algorithms. Unlike model-based RL, which learns a model of the environment but may inherit biases from simulations that do not accurately reflect real-world conditions, model-free RL directly trains policies through environmental interaction without relying on an explicit environmental model. Although model-free RL requires more computational samples and resources, it can train more robust policies that allow robots to traverse challenging environments.

Many sophisticated model-free RL algorithms exist, which can be broadly classified into two categories: policy-based (or policy optimisation) and value-based approaches. Value-based methods, e.g. Q-learning, State–Action–Reward–State–Action (SARSA), and Deep Q-learning (DQN) \cite{2021_DeepQ_DNQ_stepping_animationlonly_steppingstones}, excel only in discrete action spaces and often struggle with high-dimensional action spaces. Q-learning is an off-policy algorithm that directly learns the optimal Q-values, allowing it to derive the best possible actions irrespective of the current policy. SARSA, an on-policy variant, updates its Q-values based on the actual actions taken, making it robust in environments where the policy evolves during learning. DQN extends Q-learning by using deep NNs to approximate Q-values, enabling the algorithm to tackle complex state spaces, though it still faces challenges with high-dimensional action spaces due to difficulties in accurate value estimation. In contrast, policy-based methods, such as policy-gradient techniques, can handle complex tasks but are generally less sample-efficient than value-based methods.

More advanced algorithms combine both policy-based and value-based methods. The Actor–Critic (AC) framework simultaneously learns both a policy (actor) and a value function (critic), combining the advantages of both approaches \cite{2015_continous_control_deep_reinforcement_learning,2016_guided_learning_of_Control_Graphs_physics-based_characters}. Popular algorithms such as Trust Region Policy Optimisation (TRPO) \cite{2015_TRPO_foundamentalpaper} and PPO, based on policy-based methods, borrow ideas from AC. Moreover, other novel algorithms based on the AC framework include Deep Deterministic Policy Gradient (DDPG) \cite{2023_DDPG_e2e_openaigym_mujoco}, Twin Delayed Deep Deterministic Policy Gradients (TD3) \cite{2019_stephen_TD3_instruction}, A2C (Advantage Actor–Critic), A3C (Asynchronous Advantage Actor–Critic) \cite{2022_Jie_A3C_sim2real}, and SAC (Soft Actor–Critic) \cite{2022_Chen_transition_quadtobiped_TD3_SAC_Residual}. Each algorithm has its strengths for different tasks in bipedal locomotion scenarios. Several key factors determine their performance, such as sample efficiency, robustness and generalisation, and implementation complexity. A comparative analysis \cite{2023_Omur_comparison_RLalgorithm_bipedallocomotion} illustrates that SAC-based algorithms excel in stability and achieve the highest scores, while their training efficiency significantly trails behind that of PPO, which attains relatively high scores.

In \cite{2017_Schulman_PPO_instructure}, PPO demonstrates robustness and computational efficiency in complex scenarios such as bipedal locomotion, utilising fewer resources than TRPO. In terms of training time, PPO is much faster than SAC and DDPG \cite{2023_Omur_comparison_RLalgorithm_bipedallocomotion}. Moreover, many studies \cite{2021_siekmann_sim2real_nonreference_perodicreward_DRL_e2e_LSTM_PPO_cassie,2021_Kevin_gaitlibrary_heirarchy_reducedorder_gaitlibrary_sim2real,2020_Siekmann_drl_e2e_s2r_cassie} have demonstrated its robustness and ease of implementation. Combined with its flexibility to integrate with various NN architectures, this has made PPO the most popular choice in the field. Numerous studies have shown that PPO can enable the exploration of walking \cite{2021_siekmann_sim2real_nonreference_perodicreward_DRL_e2e_LSTM_PPO_cassie}, jumping \cite{2023_UCB_Cassie_RL_Jumping}, stair climbing \cite{2021_siekmann_blind_DRL_stair}, and stepping-stone traversal \cite{2022_OSUDRL_steppingstone_referencefree_predictionfeasiblefootsteps_camera_benchmark}, demonstrating its efficiency, robustness, and generalisation.

Additionally, the DDPG algorithm integrates the Actor–Critic framework with DQN to facilitate off-policy training, further optimising sample efficiency. In certain scenarios, such as jumping, DDPG shows higher rewards and better learning performance than PPO \cite{2023_Chongben_jumping_DDPG_multi-task_network_simulationonly,2022_Chongben_pallel_DDPG_e23_silumation_only}. TD3, developed from DDPG, improves upon the performance of both DDPG and SAC \cite{2022_Chen_transition_quadtobiped_TD3_SAC_Residual}.

SAC improves exploration through its stochastic policy and entropy-regularised objective, which encourages the agent to maintain randomness in its actions, balancing exploration and exploitation more effectively than DDPG and TD3. Unlike PPO, which is an on-policy algorithm, SAC’s off-policy nature allows it to leverage a replay buffer, reusing past experiences for training without requiring constant interaction with the environment. This, combined with entropy maximisation, enables SAC to achieve faster convergence in complex environments where exploration is essential. SAC is also known for its stability and strong performance across a wide range of tasks \cite{2022_Chen_transition_quadtobiped_TD3_SAC_Residual}. While A2C offers improved efficiency and stability compared with A3C, the asynchronous update mechanism of A3C provides better exploration capability and accelerates learning. Although these algorithms demonstrate clear advantages, they are more challenging to apply owing to their complexity compared with PPO.

\section{Training Simulation Environment} \label{Appendix:Training_Environment}

The development of DRL algorithms and sim-to-real techniques highlights the requirement for high-quality simulators. Creating a reliable simulation environment and conducting RL training is challenging. The literature shows that several simulators are available, including Isaac Gym \cite{2021_Sim_Isaac}, RoboCup3D \cite{2003_Sim_Robocup}, OpenAI Gym \cite{2016_Sim_OpenAI}, MuJoCo \cite{2012_Sim_Mujoco}, Orbit \cite{2023_Orbit_simulator}, Brax \cite{2021_Brax}, and Isaac Lab \cite{isaaclab2023}. 

OpenAI developed Gym and Gymnasium to provide lightweight environments for rapid testing of RL algorithms, including simplified bipedal locomotion models. RoboCup also serves as a benchmark platform for RL research and development in multi-agent settings.

For physics-based simulation, MuJoCo, developed by DeepMind, and Gazebo are widely used platforms that support a range of robotics research tasks. NVIDIA’s Isaac Gym, although now deprecated, played an important role as a high-performance GPU-based simulator for training agents in complex environments. Its successors, such as Isaac Lab and Orbit, continue to evolve as modern RL and robotics frameworks.

One of the most crucial aspects is the parallelisation strategy and GPU simulation. For instance, Isaac Gym was developed to maximise the throughput of physics-based machine learning algorithms, with particular emphasis on simulations requiring large numbers of environment instances executing in parallel. Running the physics simulation on a GPU can result in significant speed-ups, especially for large scenes with thousands of individual actors.

\end{appendices}


%





\ifCLASSOPTIONcaptionsoff
  \newpage
\fi



%



\bibliographystyle{IEEEtran}
\bibliography{sn-bibliography2}

@IEEEtranBSTCTL{IEEEexample:BSTcontrol,
CTLuse_forced_etal       = "yes",
CTLmax_names_forced_etal = "3",
CTLnames_show_etal       = "2" }

@INPROCEEDINGS{2004_Tedrake_RL_bipedal_policygradient,
  author={Tedrake, R. and Zhang, T.W. and Seung, H.S.},
  booktitle={IEEE/RSJ International Conference on Intelligent Robots and Systems}, 
  title={Stochastic policy gradient reinforcement learning on a simple {3D} biped}, 
  year={2004},
  pages={2849-2854},
  doi={10.1109/IROS.2004.1389841}}

@article{humphreys2024bio,
  title={Learning to Adapt through Bio-Inspired Gait Strategies for Versatile Quadruped Locomotion},
  author={Humphreys, Joseph and Zhou, Chengxu},
  journal={arXiv preprint arXiv:2412.09440},
  year={2024}
}

@inproceedings{2020_Siekmann_drl_e2e_s2r_cassie,
  author       = {Jonah Siekmann and
                  Srikar Valluri and
                  Jeremy Dao and
                  Lorenzo Bermillo and
                  Helei Duan and
                  Alan Fern and
                  Jonathan W. Hurst},
  title        = {Learning Memory-Based Control for Human-Scale Bipedal Locomotion},
  booktitle={Robotics science and systems},
  year         = {2020}
}

@article{2017_heess_emergence_animation_PPO_comprarisonrlalgorithm_differenttarrains,
author = {Heess, Nicolas and TB, Dhruva and Sriram, Srinivasan and Lemmon, Jay and Merel, Josh and Wayne, Greg and Tassa, Yuval and Erez, Tom and Wang, Ziyu and Eslami, Ali and Riedmiller, Martin and Silver, David},
year = {2017},
pages = {},
title = {Emergence of Locomotion Behaviours in Rich Environments},
journal={arXiv preprint arXiv:1707.02286}
}

@article{2017_Josh_drl_animation_imitationlearning_GRAIL_TRPO_MUJOCO,
author = {Merel, Josh and Tassa, Yuval and TB, Dhruva and Srinivasan, Sriram and Lemmon, Jay and Wang, Ziyu and Wayne, Greg and Heess, Nicolas},
year = {2017},
journal={arXiv e-prints},
pages={arXiv--1707},
title = {Learning human behaviors from motion capture by adversarial imitation}
}

@article{2017_Xuebinpeng_animation_deeploco_DRL_hiarachysystem_referencemotion_deepconvolutionalneuralnewtwork_jointangle_soccer.,
author = {Peng, Xue and Berseth, Glen and Yin, Kangkang and {van de Panne}, Michiel},
year = {2017},
pages = {1-13},
title = {{DeepLoco}: dynamic locomotion skills using hierarchical deep reinforcement learning},
volume = {36},
journal = {ACM Transactions on Graphics},
doi = {10.1145/3072959.3073602}
}

@article{2018_deepmimic_animation_PPO_deepmimic_desrieddirection_physics-based_animation_ATLAS_task,
author = {Peng, Xue and Abbeel, Pieter and Levine, Sergey and {van de Panne}, Michiel},
year = {2018},
pages = {},
title = {Deep{M}imic: Example-Guided Deep Reinforcement Learning of Physics-Based Character Skills},
volume = {37},
journal = {ACM Transactions on Graphics},
doi = {10.1145/3197517.3201311}
}

@article{2018_wenhao_yu_DRL_withoutpredefine_symmetrygait_PPO_jointanglePD_,
  title={Learning symmetric and low-energy locomotion},
  author={Yu, Wenhao and Turk, Greg and Liu, C Karen},
  journal={ACM Transactions on Graphics},
  volume={37},
  pages={1--12},
  year={2018},
  doi = {10.1145/3197517.3201397}
}

@inproceedings{2021_siekmann_sim2real_nonreference_perodicreward_DRL_e2e_LSTM_PPO_cassie,
  title={Sim-to-real learning of all common bipedal gaits via periodic reward composition},
  author={Siekmann, Jonah and Godse, Yesh and Fern, Alan and Hurst, Jonathan},
  booktitle={IEEE International Conference on Robotics and Automation},
  pages={7309--7315},
  year={2021}
}

@inproceedings{2021_duan_DRL_task-spaceaction_hiarachycotnrolscheme_inversedynamiccontroller,
  title={Learning task space actions for bipedal locomotion},
  author={Duan, Helei and Dao, Jeremy and Green, Kevin and Apgar, Taylor and Fern, Alan and Hurst, Jonathan},
  booktitle={IEEE International Conference on Robotics and Automation},
  pages={1276--1282},
  year={2021},
}

@ARTICLE{2020_motioncapture_referencebased_e2e_simulation_pybullet_augmentedrewarddesign,
  author={Yang, Chuanyu and Yuan, Kai and Heng, Shuai and Komura, Taku and Li, Zhibin},
  journal={IEEE Robotics and Automation Letters}, 
  title={Learning Natural Locomotion Behaviors for Humanoid Robots Using Human Bias}, 
  year={2020},
  volume={5},
  pages={2610-2617},
  doi={10.1109/LRA.2020.2972879}}

@article{2020_zhaoming_drl_steppingstones_PPOwithactorcritic_referencefree_simulation,
author = {Xie, Zhaoming and Ling, Hung and Kim, Nam and {van de Panne}, Michiel},
year = {2020},
pages = {213-224},
title = {{ALLSTEPS}: Curriculum‐driven Learning of Stepping Stone Skills},
volume = {39},
journal = {Computer Graphics Forum},
doi = {10.1111/cgf.14115}
}

@article{2023_wangsong_DRL_footsteptrakcing_hybridscheme_experiment_PPO_stair_task_LIPM_curriculumlearning,
  title={Learning 3{D} Bipedal Walking with Planned Footsteps and Fourier Series Periodic Gait Planning},
  author={Wang, Song and Piao, Songhao and Leng, Xiaokun and He, Zhicheng},
  journal={Sensors},
  volume={23},
  pages={1873},
  year={2023},
}

@article{2020_park_recoverstability_drl_PPO_referencebased_simulation,
  title={Understanding the stability of deep control policies for biped locomotion},
  author={Park, Hwangpil and Yu, Ri and Lee, Yoonsang and Lee, Kyungho and Lee, Jehee},
  journal={The Visual Computer},
  volume={39},
  pages={1--15},
  year={2020},
}

@inproceedings{2022_sim2real_footstepconstraint_OSUDRL_specifytouchdownlocation_actorcritic_PPO_LSTM_plannar_transisionmodel_model-predictiveplanning_CNN_predictnexttdlocation,
  title={Sim-to-real learning of footstep-constrained bipedal dynamic walking},
  author={Duan, Helei and Malik, Ashish and Dao, Jeremy and Saxena, Aseem and Green, Kevin and Siekmann, Jonah and Fern, Alan and Hurst, Jonathan},
  booktitle={International Conference on Robotics and Automation},
  pages={10428--10434},
  year={2022},
}

@INPROCEEDINGS{2022_OSUDRL_steppingstone_referencefree_predictionfeasiblefootsteps_camera_benchmark,
  author={Duan, Helei and Malik, Ashish and Gadde, Mohitvishnu S. and Dao, Jeremy and Fern, Alan and Hurst, Jonathan},
  booktitle={IEEE/RSJ International Conference on Intelligent Robots and Systems}, 
  title={Learning Dynamic Bipedal Walking Across Stepping Stones}, 
  year={2022},
  pages={6746-6752},
  doi={10.1109/IROS47612.2022.9981884}}

@article{2023_duan_OSUDRL_heightmap_visionbased_hybrid,
      title={Learning Vision-Based Bipedal Locomotion for Challenging Terrain}, 
      author = {Marum, Bart and Sabatelli, Matthia and Kasaei, Hamidreza},
      journal={arXiv preprint arXiv:2309.14594},
      year={2023},
      pages = {}
}

@article{2023_vanmarum_visionDRL_studentteacher_irregularterrain_PPO_periodicrewardfunction,
  title={Learning Perceptive Bipedal Locomotion over Irregular Terrain},
  author={van Marum, Bart and Sabatelli, Matthia and Kasaei, Hamidreza},
  journal={arXiv preprint arXiv:2304.07236},
  year={2023}
}

@article{2023_dao_OSUDRL_loco-manipulation_hybrid_simtoreal,
      title={Sim-to-Real Learning for Humanoid Box Loco-Manipulation},
  author={Dao, Jeremy and Duan, Helei and Fern, Alan},
  journal={arXiv preprint arXiv:2310.03191},
  year={2023}
}

@INPROCEEDINGS{2021_Casillo_cyberbotics_hierarchyscheme_bezierpolynominals,
  author={Castillo, Guillermo A. and Weng, Bowen and Zhang, Wei and Hereid, Ayonga},
  booktitle={IEEE/RSJ International Conference on Intelligent Robots and Systems}, 
  title={Robust Feedback Motion Policy Design Using Reinforcement Learning on a 3{D} Digit Bipedal Robot}, 
  year={2021},
  pages={5136-5143},
  doi={10.1109/IROS51168.2021.9636467}}

@INPROCEEDINGS{2023_template_taskspace_hierarchyscheme_reducedorderstateALIP_learnedhigherlevel_lowlevelinversedynamiccontroller,
  author={Castillo, Guillermo A. and Weng, Bowen and Yang, Shunpeng and Zhang, Wei and Hereid, Ayonga},
  booktitle={IEEE/RSJ International Conference on Intelligent Robots and Systems}, 
  title={Template Model Inspired Task Space Learning for Robust Bipedal Locomotion}, 
  year={2023},
  pages={8582-8589},
  doi={10.1109/IROS55552.2023.10341263}}

@INPROCEEDINGS{2021_UCB_hybridrobotics_sim2real_referencebased_HZD_gaitlibrary_e2epolicy_drl_Cassie_lowpassfilter,
  author={Li, Zhongyu and Cheng, Xuxin and Peng, Xue Bin and Abbeel, Pieter and Levine, Sergey and Berseth, Glen and Sreenath, Koushil},
  booktitle={IEEE International Conference on Robotics and Automation}, 
  title={Reinforcement Learning for Robust Parameterized Locomotion Control of Bipedal Robots}, 
  year={2021},
  pages={2811-2817},
  doi={10.1109/ICRA48506.2021.9560769}}

@INPROCEEDINGS{2022_Li_UCB_safteystability_model-based_DRL_PPO, 
    AUTHOR    = {Zhongyu Li AND Jun Zeng AND Akshay Thirugnanam AND Koushil Sreenath}, 
    TITLE     = {{Bridging Model-based Safety and Model-free Reinforcement Learning through System Identification of Low Dimensional Linear Models}}, 
    BOOKTITLE = {Proceedings of Robotics: Science and Systems}, 
    YEAR      = {2022}, 
    DOI       = {10.15607/RSS.2022.XVIII.033} 
}

@INPROCEEDINGS{ 2023_UCB_Cassie_RL_Jumping,
author={ Zhongyu Li and Xue Bin Peng and Pieter Abbeel and Sergey Levine and Glen Berseth and Koushil Sreenath },
title={ Robust and Versatile Bipedal Jumping Control through Multi-Task Reinforcement Learning },
booktitle={Robotics: Science and Systems},
pages = {},
year={ 2023 },
doi = {10.48550/arXiv.2302.09450}
}

@ARTICLE{2022_DRL_cascade_motionplanningpolicy_HZD_torso_PPOwithRNN,
  author={Castillo, Guillermo A. and Weng, Bowen and Zhang, Wei and Hereid, Ayonga},
  journal={IEEE Access}, 
  title={Reinforcement Learning-Based Cascade Motion Policy Design for Robust 3D Bipedal Locomotion}, 
  year={2022},
  volume={10},
  pages={20135-20148},
  doi={10.1109/ACCESS.2022.3151771}}

@article{2016_guided_learning_of_Control_Graphs_physics-based_characters,
title={Guided learning of control graphs for physics-based characters},
author={Liu, Libin and {van de Panne}, Michiel and Yin, KangKang},
journal={ACM Transactions on Graphics},
volume={35},
pages={1--14},
year={2016},
DOI="10.1145/2893476"
}

@INPROCEEDINGS{2022_Singh_humanoid_e2erlwithfootplannar_hierarchyscheme_PPO_simulation_HRP-5P,
  author={Singh, Rohan P. and Benallegue, Mehdi and Morisawa, Mitsuharu and Cisneros, Rafael and Kanehiro, Fumio},
  booktitle={IEEE-RAS International Conference on Humanoid Robots}, 
  title={Learning Bipedal Walking On Planned Footsteps For Humanoid Robots}, 
  year={2022},
  pages={686-693},
  doi={10.1109/Humanoids53995.2022.10000067}}

@InProceedings{2015_TRPO_foundamentalpaper,
  title = 	 {Trust Region Policy Optimization},
  author = 	 {Schulman, John and Levine, Sergey and Abbeel, Pieter and Jordan, Michael and Moritz, Philipp},
  booktitle = 	 {International Conference on Machine Learning},
  pages = 	 {1889--1897},
  year = 	 {2015}
}

@inproceedings{2015_continous_control_deep_reinforcement_learning,
  author       = {Timothy P. Lillicrap and
                  Jonathan J. Hunt and
                  Alexander Pritzel and
                  Nicolas Heess and
                  Tom Erez and
                  Yuval Tassa and
                  David Silver and
                  Daan Wierstra},
 
  title        = {Continuous control with deep reinforcement learning},
  booktitle    = {International Conference on Learning Representations},
  year         = {2016},
}

@InProceedings{2020_Xie_firstsim2real_,
  title = 	 {Learning Locomotion Skills for Cassie: Iterative Design and Sim-to-Real},
  author =       {Xie, Zhaoming and Clary, Patrick and Dao, Jeremy and Morais, Pedro and Hurst, Jonanthan and {van de Panne}, Michiel},
  booktitle = 	 {Conference on Robot Learning},
  pages = 	 {317--329},
  year = 	 {2020},
}

@inproceedings{2016_GAIL_imitation_learning,
  title={Generative Adversarial Imitation Learning},
  author={Jonathan Ho and Stefano Ermon},
  booktitle={International Conference on Neural Information Processing Systems},
  pages={4572--4580},
  year={2016}
}

@inproceedings{2019_Yu_Phy_system_identification,
  title={Sim-to-Real Transfer for Biped Locomotion},
  author={Wenhao Yu and Visak C. V. Kumar and Greg Turk and C. Karen Liu},
  booktitle={IEEE/RSJ International Conference on Intelligent Robots and Systems},
  year={2019},
  pages={3503-3510}
}

@INPROCEEDINGS{2021_DeepQ_DNQ_stepping_animationlonly_steppingstones,
  author={Meduri, Avadesh and Khadiv, Majid and Righetti, Ludovic},
  booktitle={IEEE International Conference on Robotics and Automation}, 
  title={Deep{Q} Stepper: A framework for reactive dynamic walking on uneven terrain}, 
  year={2021},
  pages={2099-2105},
  doi={10.1109/ICRA48506.2021.9562093}}

@INPROCEEDINGS{2018_xie_OSUDRL_feedback_DRL_reference-based,
  author={Xie, Zhaoming and Berseth, Glen and Clary, Patrick and Hurst, Jonathan and {van de Panne}, Michiel},
  booktitle={IEEE/RSJ International Conference on Intelligent Robots and Systems}, 
  title={Feedback Control For Cassie With Deep Reinforcement Learning}, 
  year={2018},
  pages={1241-1246},
  doi={10.1109/IROS.2018.8593722}}

@inproceedings{2018_yang_push_recover_wholebody,
title = "Learning Whole-body Motor Skills for Humanoids",
author = "Chuanyu Yang and Kai Yuan and Wolfgang Merkt and Taku Komura and Sethu Vijayakumar and Zhibin Li",
year = "2019",
doi = "10.1109/HUMANOIDS.2018.8625045",
pages = "270--276",
booktitle = "IEEE-RAS  International Conference on Humanoid Robots ",}

@INPROCEEDINGS{2004_model-based_RL_learnpoincaremap_simple_bipedrobot,
  author={Morimoto, J. and Cheng, G. and Atkeson, C.G. and Zeglin, G.},
  booktitle={IEEE International Conference on Robotics and Automation}, 
  title={A simple reinforcement learning algorithm for biped walking}, 
  year={2004},
  pages={3030-3035 Vol.3},
  doi={10.1109/ROBOT.2004.1307522}}

@INPROCEEDINGS{2019_tianyu_ATRIAS_PPO_joint0level_trajectory_learnedhighlevelpolicy,
  author={Li, Tianyu and Geyer, Hartmut and Atkeson, Christopher G. and Rai, Akshara},
  booktitle={International Conference on Robotics and Automation }, 
  title={Using Deep Reinforcement Learning to Learn High-Level Policies on the {ATRIAS} Biped}, 
  year={2019},
  pages={263-269},
  doi={10.1109/ICRA.2019.8793864}}

@article{2023_Rohan_sim2real_currentdecompose,
   title={Learning Bipedal Walking for Humanoids With Current Feedback},
   volume={11},
   DOI={10.1109/access.2023.3301175},
   journal={IEEE Access},
   author={singh, Rohan P. and Xie, Zhaoming and Gergondet, Pierre and Kanehiro, Fumio},
   year={2023},
   pages={82013–82023} }

@INPROCEEDINGS{2021_Diego_deepwalk_curriculum_learning_sim2real,
  author={Rodriguez, Diego and Behnke, Sven},
  booktitle={IEEE International Conference on Robotics and Automation}, 
  title={DeepWalk: Omnidirectional Bipedal Gait by Deep Reinforcement Learning}, 
  year={2021},
  pages={3033-3039},
  doi={10.1109/ICRA48506.2021.9561717}}

@article{2021_Mohammadreza_DRL_improveparameters_forplanner,
author = {Kasaei, Mohammadreza and Abreu, Miguel and Lau, Nuno and Pereira, Artur and Reis, Luís},
year = {2021},
pages = {103900},
title = {Robust biped locomotion using deep reinforcement learning on top of an analytical control approach},
volume = {146},
journal = {Robotics and Autonomous Systems},
doi = {10.1016/j.robot.2021.103900}
}

@article{2023_Omur_comparison_RLalgorithm_bipedallocomotion,
  title={Comparative Analysis of Reinforcement Learning Algorithms for Bipedal Robot Locomotion},
  author={Aydogmus, Omur and Yilmaz, Musa},
  journal={IEEE Access},
  year={2023},
  pages={7490--7499},
}

@article{2023_Chongben_jumping_DDPG_multi-task_network_simulationonly,
author = {Tao, Chongben and Li, Mengru and Cao, Feng and Gao, Zhen and Zhang, Zufeng},
year = {2023},
pages={2300352},
title = {A Multiobjective Collaborative Deep Reinforcement Learning Algorithm for Jumping Optimization of Bipedal Robot},
volume = {6},
journal = {Advanced Intelligent Systems},
doi = {10.1002/aisy.202300352}
}

@article{2024_li_ucb_unifiedframework_PPO_IOput_teacherstudentcompare_IOhistory_e2etraining,
      title={Reinforcement Learning for Versatile, Dynamic, and Robust Bipedal Locomotion Control}, 
      author={Zhongyu Li and Xue Bin Peng and Pieter Abbeel and Sergey Levine and Glen Berseth and Koushil Sreenath},
      year={2024},
     journal={arXiv e-prints},
     pages={arXiv--2401},
}

@ARTICLE{2023_DDPG_e2e_openaigym_mujoco,
  author={Huang, Changxin and Wang, Guangrun and Zhou, Zhibo and Zhang, Ronghui and Lin, Liang},
  journal={IEEE Transactions on Pattern Analysis and Machine Intelligence}, 
  title={Reward-Adaptive Reinforcement Learning: Dynamic Policy Gradient Optimization for Bipedal Locomotion}, 
  year={2023},
  volume={45},
  pages={7686-7695},
  doi={10.1109/TPAMI.2022.3223407}}

@ARTICLE{2022_Chen_transition_quadtobiped_TD3_SAC_Residual,
  author={Yu, Chen and Rosendo, Andre},
  journal={IEEE Robotics and Automation Letters}, 
  title={Multi-Modal Legged Locomotion Framework With Automated Residual Reinforcement Learning}, 
  year={2022},
  volume={7},
  pages={10312-10319},
  doi={10.1109/LRA.2022.3191071}}

@ARTICLE{2022_Chongben_pallel_DDPG_e23_silumation_only,
  author={Tao, Chongben and Xue, Jie and Zhang, Zufeng and Gao, Zhen},
  journal={IEEE Transactions on Circuits and Systems II: Express Briefs}, 
  title={Parallel Deep Reinforcement Learning Method for Gait Control of Biped Robot}, 
  year={2022},
  volume={69},
  pages={2802-2806},
  doi={10.1109/TCSII.2022.3145373}}

@ARTICLE{2022_Jie_A3C_sim2real,
  author={Leng, Jie and Fan, Suozhong and Tang, Jun and Mou, Haiming and Xue, Junxiao and Li, Qingdu},
  journal={IEEE Access}, 
  title={{M-A3C}: A Mean-Asynchronous Advantage Actor-Critic Reinforcement Learning Method for Real-Time Gait Planning of Biped Robot}, 
  year={2022},
  volume={10},
  pages={76523-76536},
  doi={10.1109/ACCESS.2022.3176608}}

@INPROCEEDINGS{2022_Ashish_RMA_sim2real_UCB,
  author={Kumar, Ashish and Li, Zhongyu and Zeng, Jun and Pathak, Deepak and Sreenath, Koushil and Malik, Jitendra},
  booktitle={IEEE/RSJ International Conference on Intelligent Robots and Systems}, 
  title={Adapting Rapid Motor Adaptation for Bipedal Robots}, 
  year={2022},
  pages={1161-1168},
  doi={10.1109/IROS47612.2022.9981091}}

@article{2022_Hou_drlasparametertoMPC,
  title={Deep Reinforcement Learning for Model Predictive Controller Based on Disturbed Single Rigid Body Model of Biped Robots},
  author={Landong Hou and Bin Li and Weilong Liu and Yiming Xu and Shu Yang and Xuewen Rong},
  journal={Machines},
  volume={10},
  pages={975},
  year={2022}
}

@INPROCEEDINGS{2021_michael_velocity-basedcontrol_motioncapturedata,
  author={Taylor, Michael and Bashkirov, Sergey and Rico, Javier Fernandez and Toriyama, Ike and Miyada, Naoyuki and Yanagisawa, Hideki and Ishizuka, Kensaku},
  booktitle={IEEE International Conference on Robotics and Automation}, 
  title={Learning Bipedal Robot Locomotion from Human Movement}, 
  year={2021},
  pages={2797-2803},
  doi={10.1109/ICRA48506.2021.9561591}}

@ARTICLE{2021_Kevin_gaitlibrary_heirarchy_reducedorder_gaitlibrary_sim2real,
  author={Green, Kevin and Godse, Yesh and Dao, Jeremy and Hatton, Ross L. and Fern, Alan and Hurst, Jonathan},
  journal={IEEE Robotics and Automation Letters}, 
  title={Learning Spring Mass Locomotion: Guiding Policies With a Reduced-Order Model}, 
  year={2021},
  volume={6},
  pages={3926-3932},
  doi={10.1109/LRA.2021.3066833}}

@article{2020_Javier_safereinforcementlearning,
author = {García, Javier and Shafie, Diogo},
year = {2020},
pages = {103360},
title = {Teaching a humanoid robot to walk faster through Safe Reinforcement Learning},
volume = {88},
journal = {Engineering Applications of Artificial Intelligence},
doi = {10.1016/j.engappai.2019.103360}
}

@article{2024_Li_model_based_LLDRL_hybrid_control_scheme,
    title={Agile and versatile bipedal robot tracking control through reinforcement learning},
    author={Jiayi Li and Linqi Ye and Yi Cheng and Houde Liu and Bin Liang},
    year={2024},
    journal={arXiv preprint arXiv:2404.08246},
}

@ARTICLE{2023_Wei_learned_hierarchy_framework_wheeled_bipedalrobot,
  author={Zhu, Wei and Hayashibe, Mitsuhiro},
  journal={IEEE Transactions on Industrial Electronics}, 
  title={A Hierarchical Deep Reinforcement Learning Framework With High Efficiency and Generalization for Fast and Safe Navigation}, 
  year={2023},
  volume={70},
  pages={4962-4971},
  doi={10.1109/TIE.2022.3190850}}

@INPROCEEDINGS{2016_LIPM_trajectoryoptimization,
  author={Herzog, Alexander and Schaal, Stefan and Righetti, Ludovic},
  booktitle={IEEE/RSJ International Conference on Intelligent Robots and Systems}, 
  title={Structured contact force optimization for kino-dynamic motion generation}, 
  year={2016},
  pages={2703-2710},
  doi={10.1109/IROS.2016.7759420}}

@INPROCEEDINGS{2013_3DSLIPmodel_running,
  author={Wensing, Patrick M. and Orin, David E.},
  booktitle={IEEE/RSJ International Conference on Intelligent Robots and Systems}, 
  title={High-speed humanoid running through control with a 3D-SLIP model}, 
  year={2013},
  pages={5134-5140},
  doi={10.1109/IROS.2013.6697099}}

@article{2019_simtoreal_actuatornet_quadrupedrobot,
  title={Learning agile and dynamic motor skills for legged robots},
  author={Hwangbo, Jemin and Lee, Joonho and Dosovitskiy, Alexey and Bellicoso, Dario and Tsounis, Vassilios and Koltun, Vladlen and Hutter, Marco},
  journal={Science Robotics},
  volume={4},
  pages={eaau5872},
  year={2019}
}

@ARTICLE{2018_Tsunekawa_visualnavigation_hierarchy_DDPG-sim2real_NAO,
  author={Lobos-Tsunekawa, Kenzo and Leiva, Francisco and Ruiz-del-Solar, Javier},
  journal={IEEE Robotics and Automation Letters}, 
  title={Visual Navigation for Biped Humanoid Robots Using Deep Reinforcement Learning}, 
  year={2018},
  volume={3},
  number={4},
  pages={3247-3254},
  doi={10.1109/LRA.2018.2851148}}

@inproceedings{2023_kumar_wordscontrol_LLM,
      title={Words into Action: Learning Diverse Humanoid Robot Behaviors using Language Guided Iterative Motion Refinement}, 
      author={K. Niranjan Kumar and Irfan Essa and Sehoon Ha},
      booktitle={Workshop on Language and Robot Learning: Language as Grounding},
      year={2023},
}

@article{2024_tong_advancements_review,
  title={Advancements in Humanoid Robots: A Comprehensive Review and Future Prospects},
  author={Tong, Yuchuang and Liu, Haotian and Zhang, Zhengtao},
  journal={IEEE/CAA Journal of Automatica Sinica},
  volume={11},
  pages={301--328},
  year={2024},
}

@article{2017_biped_model-based_method_review,
  title={A brief review of dynamics and control of underactuated biped robots},
  author={Gupta, Surbhi and Kumar, Amod},
  journal={Advanced Robotics},
  volume={31},
  pages={607--623},
  year={2017},
}

@ARTICLE{2020_systematic_review,
  author={Khan, Md. Al-Masrur and Khan, Md Rashed Jaowad and Tooshil, Abul and Sikder, Niloy and Mahmud, M. A. Parvez and Kouzani, Abbas Z. and Nahid, Abdullah-Al},
  journal={IEEE Access}, 
  title={A Systematic Review on Reinforcement Learning-Based Robotics Within the Last Decade}, 
  year={2020},
  volume={8},
  pages={176598-176623},
  doi={10.1109/ACCESS.2020.3027152}}

@inproceedings{2017_compare_actionspace_RL,
  title={Learning locomotion skills using deeprl: Does the choice of action space matter?},
  author={Peng, Xue Bin and {van de Panne}, Michiel},
  booktitle={ACM SIGGRAPH/Eurographics Symposium on Computer Animation},
  pages={1--13},
  year={2017}
}

@article{2021_Jenna_bipedrobot_review,
author = {Reher, Jenna and Ames, Aaron},
year = {2021},
pages = {},
title = {Dynamic Walking: Toward Agile and Efficient Bipedal Robots},
volume = {4},
journal = {Annual Review of Control, Robotics, and Autonomous Systems},
doi = {10.1146/annurev-control-071020-045021}
}

@article{2021_Justin_review_recentporgressinleggedrobot,
author = {Carpentier, Justin and Wieber, Pierre-Brice},
year = {2021},
pages={231--238},
title = {Recent Progress in Legged Robots Locomotion Control},
volume = {2},
journal = {Current Robotics Reports},
doi = {10.1007/s43154-021-00059-0}
}

@article{2017_Schulman_PPO_instructure,
      title={Proximal Policy Optimization Algorithms}, 
      author={John Schulman and Filip Wolski and Prafulla Dhariwal and Alec Radford and Oleg Klimov},
      year={2017},
      journal={arXiv e-prints},
      pages={arXiv--1707},
}

@ARTICLE{2017_DRL_survey,
  author={Arulkumaran, Kai and Deisenroth, Marc Peter and Brundage, Miles and Bharath, Anil Anthony},
  journal={IEEE Signal Processing Magazine}, 
  title={Deep Reinforcement Learning: A Brief Survey}, 
  year={2017},
  volume={34},
  pages={26-38},
  doi={10.1109/MSP.2017.2743240}}

@inproceedings{2021_siekmann_blind_DRL_stair,
      title={Blind Bipedal Stair Traversal via Sim-to-Real Reinforcement Learning}, 
      author={Jonah Siekmann and Kevin Green and John Warila and Alan Fern and Jonathan Hurst},
      booktitle={Robotics: Science and Systems},
      year={2021},
      doi= {10.15607/RSS.2021.XVII.061},
}

@article{2023_radosavovic_realworld_transformer_NNmodel,
  title={Real-world humanoid locomotion with reinforcement learning},
  author={Radosavovic, Ilija and Xiao, Tete and Zhang, Bike and Darrell, Trevor and Malik, Jitendra and Sreenath, Koushil},       
    journal={Science Robotics},
  volume={9},
  number={89},
  pages={eadi9579},
  year={2024},
}

@article{quadModelBased1,
  title={Demonstrating a walk in the park: Learning to walk in 20 minutes with model-free reinforcement learning},
  author={Smith, Laura and Kostrikov, Ilya and Levine, Sergey},
  journal={Robotics: Science and Systems Demo},
  volume={2},
  pages={4},
  year={2023}
}

@inproceedings{quadModelBased2,
title={Day{D}reamer: World Models for Physical Robot Learning},
author={Philipp Wu and Alejandro Escontrela and Danijar Hafner and Pieter Abbeel and Ken Goldberg},
booktitle={Conference on Robot Learning},
pages={2226--2240},
year={2023},
}

@article{quadDeformable,
author = {Suyoung Choi  and Gwanghyeon Ji  and Jeongsoo Park  and Hyeongjun Kim  and Juhyeok Mun  and Jeong Hyun Lee  and Jemin Hwangbo },
title = {Learning quadrupedal locomotion on deformable terrain},
journal = {Science Robotics},
volume = {8},
pages = {eade2256},
year = {2023},
doi = {10.1126/scirobotics.ade2256},
}

@InProceedings{quadModular,
  title = 	 {GenLoco: Generalized Locomotion Controllers for Quadrupedal Robots},
  author =       {Feng, Gilbert and Zhang, Hongbo and Li, Zhongyu and Peng, Xue Bin and Basireddy, Bhuvan and Yue, Linzhu and SONG, ZHITAO and Yang, Lizhi and Liu, Yunhui and Sreenath, Koushil and Levine, Sergey},
  booktitle = 	 {Conference on Robot Learning},
  pages = 	 {1893--1903},
  year = 	 {2023},
  volume = 	 {205},
}

@INPROCEEDINGS{quadImitation2,
  author={Fuchioka, Yuni and Xie, Zhaoming and {van de Panne}, Michiel},
  booktitle={IEEE International Conference on Robotics and Automation}, 
  title={{OPT}-{M}imic: Imitation of Optimized Trajectories for Dynamic Quadruped Behaviors}, 
  year={2023},
  pages={5092-5098},
  doi={10.1109/ICRA48891.2023.10160562}
}

@ARTICLE{quadRLPlanning,
  author={Gangapurwala, Siddhant and Geisert, Mathieu and Orsolino, Romeo and Fallon, Maurice and Havoutis, Ioannis},
  journal={IEEE Transactions on Robotics}, 
  title={{RLOC}: Terrain-Aware Legged Locomotion Using Reinforcement Learning and Optimal Control}, 
  year={2022},
  volume={38},
  pages={2908-2927},
  doi={10.1109/TRO.2022.3172469}
}

@article{quadRLLL,
author = {Fabian Jenelten  and Junzhe He  and Farbod Farshidian  and Marco Hutter },
title = {{DTC}: Deep Tracking Control},
journal = {Science Robotics},
volume = {9},
pages = {eadh5401},
year = {2024},
doi = {10.1126/scirobotics.adh5401},
}

@inproceedings{quadAnimal1,
	author = {Peng, Xue Bin and Coumans, Erwin and Zhang, Tingnan and Lee, Tsang-Wei Edward and Tan, Jie and Levine, Sergey},
	booktitle={Robotics: Science and Systems},
	year = {2020},
	title = {Learning Agile Robotic Locomotion Skills by Imitating Animals},
	doi = {10.15607/RSS.2020.XVI.064}
}

@INPROCEEDINGS{quadAnimal2,
  author={Yin, Fulong and Tang, Annan and Xu, Liangwei and Cao, Yue and Zheng, Yu and Zhang, Zhengyou and Chen, Xiangyu},
  booktitle={IEEE/RSJ International Conference on Intelligent Robots and Systems}, 
  title={Run Like a Dog: Learning Based Whole-Body Control Framework for Quadruped Gait Style Transfer}, 
  year={2021},
  pages={8508-8514},
  doi={10.1109/IROS51168.2021.9636805}
}

@ARTICLE{quadManip1,
  author={Ma, Yuntao and Farshidian, Farbod and Miki, Takahiro and Lee, Joonho and Hutter, Marco},
  journal={IEEE Robotics and Automation Letters}, 
  title={Combining Learning-Based Locomotion Policy With Model-Based Manipulation for Legged Mobile Manipulators}, 
  year={2022},
  volume={7},
  pages={2377-2384},
  doi={10.1109/LRA.2022.3143567}
}

@inproceedings{quadManip2,
title={Deep Whole-Body Control: Learning a Unified Policy for Manipulation and Locomotion},
author={Zipeng Fu and Xuxin Cheng and Deepak Pathak},
booktitle={Conference on Robot Learning},
pages={138--149},
year={2023},
}

@inproceedings{quadManip3,
title={Pedipulate: Enabling Manipulation Skills using a Quadruped Robot’s Leg},
author={Arm, Philip and Mittal, Mayank and Kolvenbach, Hendrik and Hutter, Marco},
booktitle={IEEE Conference on Robotics and Automation},
year={2024}
}

@ARTICLE{quadSpace1,
  author={Rudin, Nikita and Kolvenbach, Hendrik and Tsounis, Vassilios and Hutter, Marco},
  journal={IEEE Transactions on Robotics}, 
  title={Cat-Like Jumping and Landing of Legged Robots in Low Gravity Using Deep Reinforcement Learning}, 
  year={2022},
  volume={38},
  pages={317-328},
  doi={10.1109/TRO.2021.3084374}
}

@article{quadSpace2,
  title={Reinforcement learning-based stable jump control method for asteroid-exploration quadruped robots},
  author={Qi, Ji and Gao, Haibo and Su, Huanli and Han, Liangliang and Su, Bo and Huo, Mingying and Yu, Haitao and Deng, Zongquan},
  journal={Aerospace Science and Technology},
  volume={142},
  pages={108689},
  year={2023},
}

@ARTICLE{TAMOLS,
  author={Jenelten, Fabian and Grandia, Ruben and Farshidian, Farbod and Hutter, Marco},
  journal={IEEE Transactions on Robotics}, 
  title={{TAMOLS}: Terrain-Aware Motion Optimization for Legged Systems}, 
  year={2022},
  volume={38},
  pages={3395-3413},
  doi={10.1109/TRO.2022.3186804}
}

@ARTICLE{quadmpcref,
  author={Kang, Dongho and Cheng, Jin and Zamora, Miguel and Zargarbashi, Fatemeh and Coros, Stelian},
  journal={IEEE Robotics and Automation Letters}, 
  title={{RL + Model-Based Control}: Using On-Demand Optimal Control to Learn Versatile Legged Locomotion}, 
  year={2023},
  volume={8},
  pages={6619-6626},
  doi={10.1109/LRA.2023.3307008}
}

@book{2003_Sim_Robocup,
  title={{RoboCup} 2001: Robot Soccer World Cup V},
  author={Birk, Andreas and Coradeschi, Silvia and Tadokoro, Satoshi},
  volume={2377},
  year={2003},
}

@inproceedings{2021_Sim_Isaac,
  title={Learning to walk in minutes using massively parallel deep reinforcement learning},
  author={Rudin, Nikita and Hoeller, David and Reist, Philipp and Hutter, Marco},
  booktitle={Conference on Robot Learning},
  pages={91--100},
  year={2022},
}

@article{2016_Sim_OpenAI,
  title={Openai gym},
  author={Brockman, Greg and Cheung, Vicki and Pettersson, Ludwig and Schneider, Jonas and Schulman, John and Tang, Jie and Zaremba, Wojciech},
  journal={arXiv preprint arXiv:1606.01540},
  year={2016}
}

@inproceedings{2012_Sim_Mujoco,
  title={{MuJoCo}: A physics engine for model-based control},
  author={Todorov, Emanuel and Erez, Tom and Tassa, Yuval},
  booktitle={IEEE/RSJ International conference on intelligent robots and systems},
  pages={5026--5033},
  year={2012},
}

@ARTICLE{2023_junheng_steppingstones_MPCWBC,
  author={Li, Junheng and Nguyen, Quan},
  journal={IEEE Control Systems Letters}, 
  title={Dynamic Walking of Bipedal Robots on Uneven Stepping Stones via Adaptive-Frequency MPC}, 
  year={2023},
  volume={7},
  pages={1279-1284},
  doi={10.1109/LCSYS.2023.3234769}}

@INPROCEEDINGS{2016_Quan_gaitlibrary,
  author={Nguyen, Quan and Hereid, Ayonga and Grizzle, Jessy W. and Ames, Aaron D. and Sreenath, Koushil},
  booktitle={IEEE Conference on Decision and Control}, 
  title={3D dynamic walking on stepping stones with control barrier functions}, 
  year={2016},
  pages={827-834},
  doi={10.1109/CDC.2016.7798370}}

@misc{2023_vidoeo_demo_digit_locomaniputlation,
title = "{6+ Hours Live Autonomous Robot Demo}",
howpublished = {\url{https://www.youtube.com/watch?v=Ke468Mv8ldM}},
month = "Mar.",
year = "2024",}

@article{2019_bingjing_human_exoskeleton,
  title={Human--robot interactive control based on reinforcement learning for gait rehabilitation training robot},
  author={Bingjing, Guo and Jianhai, Han and Xiangpan, Li and Lin, Yan},
  journal={International Journal of Advanced Robotic Systems},
  volume={16},
  pages={1729881419839584},
  year={2019},
}

@inproceedings{2019_stephen_TD3_instruction,
  title={Twin-Delayed {DDPG}: A Deep Reinforcement Learning Technique to Model a Continuous Movement of an Intelligent Robot Agent},
  author={Dankwa, Stephen and Zheng, Wenfeng},
  booktitle={International conference on vision, image and signal processing},
  pages={1--5},
  year={2019}
}

@article{2024_Zhang_wholebody_adversarial_motion_priors,
      title={Whole-body Humanoid Robot Locomotion with Human Reference}, 
      author={Qiang Zhang and Peter Cui and David Yan and Jingkai Sun and Yiqun Duan and Arthur Zhang and Renjing Xu},
      journal={arXiv preprint arXiv:2402.18294},
      year={2024}
}

@article{2024_tang_humanmimic_adversarial,
      title={{HumanMimic}: Learning Natural Locomotion and Transitions for Humanoid Robot via Wasserstein Adversarial Imitation}, 
      author={Annan Tang and Takuma Hiraoka and Naoki Hiraoka and Fan Shi and Kento Kawaharazuka and Kunio Kojima and Kei Okada and Masayuki Inaba},
      journal={arXiv preprint arXiv:2309.14225},
      year={2023}
}

@article{2024_gaspard_footstepnet_soccer,
      title={{FootstepNet}: an Efficient Actor-Critic Method for Fast On-line Bipedal Footstep Planning and Forecasting}, 
      author={Clément Gaspard and Grégoire Passault and Mélodie Daniel and Olivier Ly},
      journal={arXiv preprint arXiv:2403.12589},
      year={2024}
}

@article{2024_speratebody_expressive_allmotion,
      title={Expressive Whole-Body Control for Humanoid Robots}, 
      author={Xuxin Cheng and Yandong Ji and Junming Chen and Ruihan Yang and Ge Yang and Xiaolong Wang},
      journal={arXiv preprint arXiv:2402.16796},
      year={2024}
}

@inproceedings{2023_seo_deep_loco_manipulation,
title={Deep Imitation Learning for Humanoid Loco-manipulation through Human Teleoperation},
author={Seo, Mingyo and Han, Steve and Sim, Kyutae and Bang, Seung Hyeon and Gonzalez, Carlos and Sentis, Luis and Zhu, Yuke},
booktitle={IEEE-RAS International Conference on Humanoid Robots},
pages={1--8},
year={2023},
}

@inproceedings{2023_masuda_simtoreal_identification,
title={Sim-to-real transfer of compliant bipedal locomotion on torque sensor-less gear-driven humanoid},
author={Masuda, Shimpei and Takahashi, Kuniyuki},
booktitle={IEEE-RAS International Conference on Humanoid Robots},
pages={1--8},
year={2023},
}

@article{2023_deepmind_haarnoja_playsoccer,
title={Learning agile soccer skills for a bipedal robot with deep reinforcement learning},
author={Haarnoja, Tuomas and Moran, Ben and Lever, Guy and Huang, Sandy H and Tirumala, Dhruva and Humplik, Jan and Wulfmeier, Markus and Tunyasuvunakool, Saran and Siegel, Noah Y and Hafner, Roland and others},
journal={Science Robotics},
volume={9},
pages={eadi8022},
year={2024},
}

@article{2023_baltes_control_scooter,
  title={A deep reinforcement learning algorithm to control a two-wheeled scooter with a humanoid robot},
  author={Baltes, Jacky and Christmann, Guilherme and Saeedvand, Saeed},
  journal={Engineering Applications of Artificial Intelligence},
  volume={126},
  pages={106941},
  year={2023},
}

@inproceedings{2022_byravan_nerf2real_basketball,
title={{Nerf2real}: {Sim2real} transfer of vision-guided bipedal motion skills using neural radiance fields},
author={Byravan, Arunkumar and Humplik, Jan and Hasenclever, Leonard and Brussee, Arthur and Nori, Francesco and Haarnoja, Tuomas and Moran, Ben and Bohez, Steven and Sadeghi, Fereshteh and Vujatovic, Bojan and others},
booktitle={IEEE International Conference on Robotics and Automation},
pages={9362--9369},
year={2023},
}

@article{industrial_robotics_applications,
  title={Advanced applications of industrial robotics: New trends and possibilities},
  author={Dzedzickis, Andrius and Suba{\v{c}}i{\=u}t{\.e}-{\v{Z}}emaitien{\.e}, Jurga and {\v{S}}utinys, Ernestas and Samukait{\.e}-Bubnien{\.e}, Urt{\.e} and Bu{\v{c}}inskas, Vytautas},
  journal={Applied Sciences},
  volume={12},
  pages={135},
  year={2021},
}

@article{2015_underwaterrobot_review,
  title={Underwater robots: a review of technologies and applications},
  author={Bogue, Robert},
  journal={Industrial Robot: An International Journal},
  volume={42},
  pages={186--191},
  year={2015},
}

@inproceedings{collaborative_industrial,
  title={Collaborative mobile industrial manipulator: a review of system architecture and applications},
  author={Yang, Manman and Yang, Erfu and Zante, Remi Christophe and Post, Mark and Liu, Xuefeng},
  booktitle={International conference on automation and computing },
  pages={1--6},
  year={2019},
}

@article{2024_stanford_humanplus_hierarchical_humanoidshadowingimitation,
      title={HumanPlus: Humanoid Shadowing and Imitation from Humans}, 
      author={Zipeng Fu and Qingqing Zhao and Qi Wu and Gordon Wetzstein and Chelsea Finn},
      journal = {arXiv preprint arXiv:2406.10454},
      year={2024},
      url={https://arxiv.org/abs/2406.10454}, 
}

@article{2024_LLM_AnyBipe,
       title={AnyBipe: An End-to-End Framework for Training and Deploying Bipedal Robots Guided by Large Language Models}, 
       author={Yifei Yao and Wentao He and Chenyu Gu and Jiaheng Du and Fuwei Tan and Zhen Zhu and Junguo Lu},
       year={2024},
       eprint={2409.08904},
       archivePrefix={arXiv},
       primaryClass={cs.RO},
       url={https://arxiv.org/abs/2409.08904}, 
 }

@article{2024_VLM_PC,
  title={Commonsense Reasoning for Legged Robot Adaptation with Vision-Language Models},
  author={Chen, Annie S and Lessing, Alec M and Tang, Andy and Chada, Govind and Smith, Laura and Levine, Sergey and Finn, Chelsea},
  journal={arXiv preprint arXiv:2407.02666},
  year={2024}
}

@InProceedings{2023_VLA_Rt-2,
  title = 	 {RT-2: Vision-Language-Action Models Transfer Web Knowledge to Robotic Control},
  author =       {Zitkovich, Brianna and Yu, Tianhe and Xu, Sichun and Xu, Peng and Xiao, Ted and Xia, Fei and Wu, Jialin and Wohlhart, Paul and Welker, Stefan and Wahid, Ayzaan and Vuong, Quan and Vanhoucke, Vincent and Tran, Huong and Soricut, Radu and Singh, Anikait and Singh, Jaspiar and Sermanet, Pierre and Sanketi, Pannag R. and Salazar, Grecia and Ryoo, Michael S. and Reymann, Krista and Rao, Kanishka and Pertsch, Karl and Mordatch, Igor and Michalewski, Henryk and Lu, Yao and Levine, Sergey and Lee, Lisa and Lee, Tsang-Wei Edward and Leal, Isabel and Kuang, Yuheng and Kalashnikov, Dmitry and Julian, Ryan and Joshi, Nikhil J. and Irpan, Alex and Ichter, Brian and Hsu, Jasmine and Herzog, Alexander and Hausman, Karol and Gopalakrishnan, Keerthana and Fu, Chuyuan and Florence, Pete and Finn, Chelsea and Dubey, Kumar Avinava and Driess, Danny and Ding, Tianli and Choromanski, Krzysztof Marcin and Chen, Xi and Chebotar, Yevgen and Carbajal, Justice and Brown, Noah and Brohan, Anthony and Arenas, Montserrat Gonzalez and Han, Kehang},
  booktitle = 	 {Proceedings of The 7th Conference on Robot Learning},
  pages = 	 {2165--2183},
  year = 	 {2023},
  editor = 	 {Tan, Jie and Toussaint, Marc and Darvish, Kourosh},
  url = 	 {https://proceedings.mlr.press/v229/zitkovich23a.html},
}

@inproceedings{2019_Naureen_AMASS,
  author={Mahmood, Naureen and Ghorbani, Nima and Troje, Nikolaus F. and Pons-Moll, Gerard and Black, Michael},
  booktitle={2019 IEEE/CVF International Conference on Computer Vision}, 
  title={AMASS: Archive of Motion Capture As Surface Shapes}, 
  year={2019},
  pages={5441-5450},
  doi={10.1109/ICCV.2019.00554}}

@article{2024_radosavovic_humanoid_locomotiontokenprediction_transformer,
      title={Humanoid Locomotion as Next Token Prediction}, 
      author={Ilija Radosavovic and Bike Zhang and Baifeng Shi and Jathushan Rajasegaran and Sarthak Kamat and Trevor Darrell and Koushil Sreenath and Jitendra Malik},
      year={2024},
      journal={arXiv preprint arXiv:2402.19469},
      url={https://arxiv.org/abs/2402.19469}, 
}

@article{2023_Fms_survey_robotics_Roya_applicationsChallengeandFuture,
  title={Foundation models in robotics: Applications, challenges, and the future},
  author={Firoozi, Roya and Tucker, Johnathan and Tian, Stephen and Majumdar, Anirudha and Sun, Jiankai and Liu, Weiyu and Zhu, Yuke and Song, Shuran and Kapoor, Ashish and Hausman, Karol and others},
  journal={The International Journal of Robotics Research},
  pages={02783649241281508},
  year={2023},

}

@article{2023_generalpurposerobot_FMs,
  title={Toward general-purpose robots via foundation models: A survey and meta-analysis},
  author={Hu, Yafei and Xie, Quanting and Jain, Vidhi and Francis, Jonathan and Patrikar, Jay and Keetha, Nikhil and Kim, Seungchan and Xie, Yaqi and Zhang, Tianyi and Fang, Hao-Shu and others},
  journal={arXiv preprint arXiv:2312.08782},
  year={2023}
}

@article{2024_motion_X_datasets,
      title={Motion-X: A Large-scale 3D Expressive Whole-body Human Motion Dataset}, 
    journal={NeurIPS},
    author={Jing Lin and Ailing Zeng and Shunlin Lu and Yuanhao Cai and Ruimao Zhang and Haoqian Wang and Lei Zhang},
      year={2024},
}

@article{2024_yangOu_LLM_hierarchical_quadrupedrobot,
  title={Long-horizon Locomotion and Manipulation on a Quadrupedal Robot with Large Language Models},
  author={Ouyang, Yutao and Li, Jinhan and Li, Yunfei and Li, Zhongyu and Yu, Chao and Sreenath, Koushil and Wu, Yi},
  journal={arXiv preprint arXiv:2404.05291},
  year={2024}
}

@inproceedings{2015_DAPRA_Challenge_1,
  title={No falls, no resets: Reliable humanoid behavior in the DARPA robotics challenge},
  author={Atkeson, Christopher G and Babu, Benzun P Wisely and Banerjee, Nandan and Berenson, Dmitry and Bove, Christoper P and Cui, Xiongyi and DeDonato, Mathew and Du, Ruixiang and Feng, Siyuan and Franklin, Perry and others},
  booktitle={IEEE-RAS 15th International Conference on Humanoid Robot},
  pages={623--630},
  year={2015}}

@inproceedings{2023_RL_finetune_RMFs,
author = {Bauer, Jakob and Baumli, Kate and Behbahani, Feryal and Bhoopchand, Avishkar and Bradley-Schmieg, Nathalie and Chang, Michael and Clay, Natalie and Collister, Adrian and Dasagi, Vibhavari and Gonzalez, Lucy and Gregor, Karol and Hughes, Edward and Kashem, Sheleem and Loks-Thompson, Maria and Openshaw, Hannah and Parker-Holder, Jack and Pathak, Shreya and Perez-Nieves, Nicolas and Rakicevic, Nemanja and Rockt\"{a}schel, Tim and Schroecker, Yannick and Singh, Satinder and Sygnowski, Jakub and Tuyls, Karl and York, Sarah and Zacherl, Alexander and Zhang, Lei},
title = {Human-timescale adaptation in an open-ended task space},
year = {2023},
booktitle = {Proceedings of the 40th International Conference on Machine Learning},
numpages = {49},
}

@INPROCEEDINGS{2019_iit_wholebody_retargeting,
  author={Penco, L. and Clement, B. and Modugno, V. and Mingo Hoffman, E. and Nava, G. and Pucci, D. and Tsagarakis, Nikos G. and Mouret, J. -B. and Ivaldi, S.},
  booktitle={2018 IEEE-RAS 18th International Conference on Humanoid Robots (Humanoids)}, 
  title={Robust Real-Time Whole-Body Motion Retargeting from Human to Humanoid}, 
  year={2018},
  volume={},
  number={},
  pages={425-432},
  doi={10.1109/HUMANOIDS.2018.8624943}}

@ARTICLE{2017_Ayusawa_motionretarget_optimization,
  author={Ayusawa, Ko and Yoshida, Eiichi},
  journal={IEEE Transactions on Robotics}, 
  title={Motion Retargeting for Humanoid Robots Based on Simultaneous Morphing Parameter Identification and Motion Optimization}, 
  year={2017},
  volume={33},
  number={6},
  pages={1343-1357},
  doi={10.1109/TRO.2017.2752711}}

@InProceedings{2024_MOTO_RLfinetune_foundationmodel,
  title = 	 {{MOTO}: Offline Pre-training to Online Fine-tuning for Model-based Robot Learning},
  author =       {Rafailov, Rafael and Hatch, Kyle Beltran and Kolev, Victor and Martin, John D. and Phielipp, Mariano and Finn, Chelsea},
  booktitle = 	 {Proceedings of The 7th Conference on Robot Learning},
  pages = 	 {3654--3671},
  year = 	 {2023},
  volume = 	 {229},
  series = 	 {Proceedings of Machine Learning Research},

}

@article{
        2024_FLaRe_RLfinetune_foundationmodel,
        title={FLaRe: Achieving Masterful and Adaptive Robot Policies with Large-Scale Reinforcement Learning Fine-Tuning},
        author={Jiaheng Hu and Rose Hendrix and Ali Farhadi and Aniruddha Kembhavi and Roberto Martin-Martin and Peter Stone and Kuo-Hao Zeng and Kiana Ehsani},
        journal={arXiv},
        year={2024},
        eprint={2409.16578},
}

@inproceedings{2022_codeaspolicy_google,
    title={Code as Policies: Language Model Programs for Embodied Control},
    author={Jacky Liang and Wenlong Huang and Fei Xia and Peng Xu and Karol Hausman and Brian Ichter and Pete Florence and Andy Zeng},
    booktitle={arXiv preprint arXiv:2209.07753},
    year={2022}
}

@article{2022_DoAsICan,
  title={Do as i can, not as i say: Grounding language in robotic affordances},
  author={Irpan, Alex and Herzog, Alexander and Toshev, Alexander Toshkov and Zeng, Andy and Brohan, Anthony and Ichter, Brian Andrew and David, Byron and Parada, Carolina and Finn, Chelsea and Tan, Clayton and others},
  journal={arXiv preprint arXiv:2204.01691},
  year={2022}
}

@ARTICLE{2023_Orbit_simulator,
  author={Mittal, Mayank and Yu, Calvin and Yu, Qinxi and Liu, Jingzhou and Rudin, Nikita and Hoeller, David and Yuan, Jia Lin and Singh, Ritvik and Guo, Yunrong and Mazhar, Hammad and Mandlekar, Ajay and Babich, Buck and State, Gavriel and Hutter, Marco and Garg, Animesh},
  journal={IEEE Robotics and Automation Letters}, 
  title={Orbit: A Unified Simulation Framework for Interactive Robot Learning Environments}, 
  year={2023},
  volume={8},
  number={6},
  pages={3740-3747},
  doi={10.1109/LRA.2023.3270034}}

@article{2021_Brax,
      title={Brax -- A Differentiable Physics Engine for Large Scale Rigid Body Simulation}, 
      author={C. Daniel Freeman and Erik Frey and Anton Raichuk and Sertan Girgin and Igor Mordatch and Olivier Bachem},
      year={2021},
      eprint={2106.13281},
      archivePrefix={arXiv},
      primaryClass={cs.RO},
      url={https://arxiv.org/abs/2106.13281}, 
}

@article{isaaclab2023,
  title={Isaac Lab: A GPU-Accelerated Simulation Framework for Multi-Modal Robot Learning},
  author={Mayank Mittal and Pascal Roth and James Tigue and Antoine Richard and Octi Zhang and Peter Du and Antonio Serrano-Muñoz and Xinjie Yao and René Zurbrügg and Nikita Rudin and Lukasz Wawrzyniak and Milad Rakhsha and Alain Denzler and Eric Heiden and Ales Borovicka and Ossama Ahmed and Iretiayo Akinola and Abrar Anwar and Mark T. Carlson and Ji Yuan Feng and Animesh Garg and Renato Gasoto and Lionel Gulich and Yijie Guo and M. Gussert and Alex Hansen and Mihir Kulkarni and Chenran Li and Wei Liu and Viktor Makoviychuk and Grzegorz Malczyk and Hammad Mazhar and Masoud Moghani and Adithyavairavan Murali and Michael Noseworthy and Alexander Poddubny and Nathan Ratliff and Welf Rehberg and Clemens Schwarke and Ritvik Singh and James Latham Smith and Bingjie Tang and Ruchik Thaker and Matthew Trepte and Karl Van Wyk and Fangzhou Yu and Alex Millane and Vikram Ramasamy and Remo Steiner and Sangeeta Subramanian and Clemens Volk and CY Chen and Neel Jawale and Ashwin Varghese Kuruttukulam and Michael A. Lin and Ajay Mandlekar and Karsten Patzwaldt and John Welsh and Huihua Zhao and Fatima Anes and Jean-Francois Lafleche and Nicolas Moënne-Loccoz and Soowan Park and Rob Stepinski and Dirk Van Gelder and Chris Amevor and Jan Carius and Jumyung Chang and Anka He Chen and Pablo de Heras Ciechomski and Gilles Daviet and Mohammad Mohajerani and Julia von Muralt and Viktor Reutskyy and Michael Sauter and Simon Schirm and Eric L. Shi and Pierre Terdiman and Kenny Vilella and Tobias Widmer and Gordon Yeoman and Tiffany Chen and Sergey Grizan and Cathy Li and Lotus Li and Connor Smith and Rafael Wiltz and Kostas Alexis and Yan Chang and David Chu and Linxi "Jim" Fan and Farbod Farshidian and Ankur Handa and Spencer Huang and Marco Hutter and Yashraj Narang and Soha Pouya and Shiwei Sheng and Yuke Zhu and Miles Macklin and Adam Moravanszky and Philipp Reist and Yunrong Guo and David Hoeller and Gavriel State},
  journal={arXiv preprint arXiv:2511.04831},
  year={2025},
  url={https://arxiv.org/abs/2511.04831}
}

@article{huang2024diffuseloco,
  title={DiffuseLoco: Real-Time Legged Locomotion Control with Diffusion from Offline Datasets},
  author={Huang, Xiaoyu and Chi, Yufeng and Wang, Ruofeng and Li, Zhongyu and Peng, Xue Bin and Shao, Sophia and Nikolic, Borivoje and Sreenath, Koushil},
  journal={arXiv preprint arXiv:2404.19264},
  year={2024}
}

%








\end{document}